%% file: main.tex
\documentclass[10pt,twocolumn,letterpaper]{article}

\usepackage[pagenumbers]{cvpr}
\input{preamble}
\definecolor{cvprblue}{rgb}{0.21,0.49,0.74}
\usepackage[pagebackref,breaklinks,colorlinks,allcolors=cvprblue]{hyperref}

\def\paperID{*****} 
\def\confName{CVPR}
\def\confYear{2026}

\title{Part-Aware Open-Vocabulary 3D Affordance Grounding via Prototypical Semantic and Geometric Alignment}

\author{
Dongqiang Gou \\
ShanghaiTech University \\
{\tt\small goudq2023@shanghaitech.edu.cn}
\and
Xuming He \\
ShanghaiTech University \\
{\tt\small hexm@shanghaitech.edu.cn}
}

\usepackage{booktabs}      
\usepackage{caption}       
\usepackage{amsmath,amssymb} 
\usepackage{graphicx}      
\usepackage{multirow}      
\usepackage{tabularx}      
\usepackage{makecell}      
\usepackage{tikz}          
\usepackage{subcaption}    
\usepackage{diagbox}       
\usepackage{rotating}      
\usepackage{colortbl}      
\usepackage{pifont}        
\newcommand{\cmark}{\ding{51}}  
\newcommand{\xmark}{\ding{55}}  

\usepackage{threeparttable}
\usepackage{siunitx}       
\usepackage{etoolbox}      
\usepackage{arydshln}      
\usepackage[table]{xcolor}
\newcommand{\g}[1]{\cellcolor{gray!10}{#1}}

\newcommand{\NA}{\makebox[2.5em]{—}}
\newcommand{\dash}{\makebox[2.5em]{---}}

\begin{document}
\maketitle
\input{sec/0_abstract}    
\input{sec/1_intro}
\input{sec/2_related_works}
\input{sec/3_method}

\input{sec/4_training}

\input{sec/5_experiments}
\input{sec/6_conclusion}

{
    \small
    \bibliographystyle{ieeenat_fullname}
    \bibliography{main}
}

\input{sec/X_suppl}


\end{document}

%% file: sec/0_abstract.tex
\begin{abstract}
Grounding natural language questions to functionally relevant regions in 3D objects—termed language-driven 3D affordance grounding—is essential for embodied intelligence and human--AI interaction. Existing methods, while progressing from label-based to language-driven approaches, still face challenges in open-vocabulary generalization, fine-grained geometric alignment, and part-level semantic consistency. To address these issues, we propose a novel two-stage cross-modal framework that enhances both semantic and geometric representations for open-vocabulary 3D affordance grounding. In the first stage, large language models generate part-aware instructions to recover missing semantics, enabling the model to link semantically similar affordances. In the second stage, we introduce two key components: Affordance Prototype Aggregation (APA), which captures cross-object geometric consistency for each affordance, and Intra-Object Relational Modeling (IORM), which refines geometric differentiation within objects to support precise semantic alignment. We validate the effectiveness of our method through extensive experiments on a newly introduced benchmark, as well as two existing benchmarks, demonstrating superior performance in comparison with existing methods.

\end{abstract}

%% file: sec/1_intro.tex
\section{Introduction}
\label{sec:intro}

The term “affordance” was introduced by J.~Gibson~\cite{Gibson_2014} as “opportunities for interaction”. Grounding natural language questions to functionally relevant regions in 3D objects—known as \textit{language-driven 3D affordance grounding}—requires understanding human intent from the question (e.g., ``Where should you grasp the bag?'') and predicting a per-point mask over the object to localize the corresponding functional area. This capability is central to applications in augmented reality and embodied AI~\cite{Huang2023VLMaps,Yamazaki2023OpenFusion,Shan2023Trans4Trans,Driess2023PaLM-E,Gadre2022Ego4D}, where agents must act upon language questions in complex 3D environments.

\begin{figure}[t]
    \centering
    \includegraphics[width=1\linewidth]{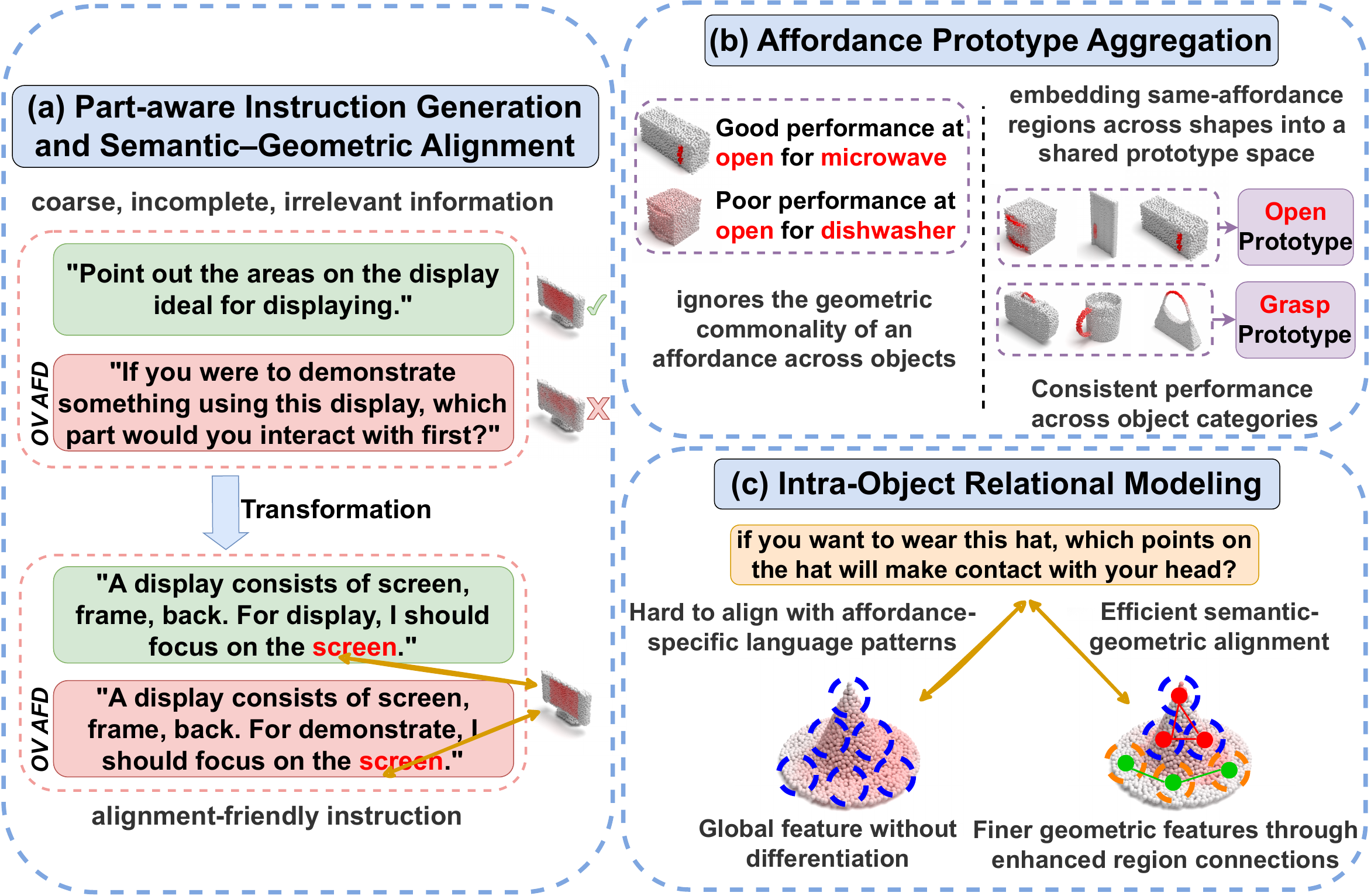}
    \caption{Overview of motivations, illustrating (a) part-aware instruction generation and semantic–geometric alignment, (b) affordance prototype aggregation, (c) intra-object relational modeling.}
    \label{fig:motivation}
\end{figure}

Existing methods for 3D affordance grounding have evolved through several stages. Early approaches, including label-based methods such as OpenAD~\cite{Nguyen2023OpenAD} and image-guided methods like IAG~\cite{Yang2023IAG} and MIFAG~\cite{Gao2025MIFAG}, either rely on predefined labels or 2D interaction cues, and thus remain limited because real-world intent is typically expressed through language.
Subsequently, language-driven models such as LASO~\cite{Li2024LASO} and Seq-AFD~\cite{Yu2025SeqAfford} grounded interactions directly from language onto 3D shapes. However, they still struggle with open-vocabulary generalization, limiting their ability to handle unseen affordances.
More recently, models like 3D-AffordanceLLM~\cite{Chu2025AffordanceLLM} have begun exploring open-vocabulary grounding, they perform alignment on coarse textual inputs often lack object-part information—crucial for understanding functional attributes—and may contain irrelevant details. This results in inefficient semantic–geometric alignment and weak open-vocabulary generalization, as the model cannot associate unseen affordances with learned ones. For instance, although the unseen affordance \textit{demonstrate} and the seen affordance \textit{display} convey highly similar functional intents, existing models fail to leverage this semantic proximity because, without part-level cues, they cannot recognize that both should correspond to the same region (Fig.~\ref{fig:motivation}(a)).
Moreover, these methods face two key limitations.
First, prior methods trained with an object–affordance–ground truth mask (obj–afd–gt mask) tend to learn a direct mapping between each object and its corresponding affordance, ignoring the geometric consistency shared by the same affordance across different objects. As a result, a model may correctly learn open on a microwave but fail on a dishwasher (Fig.~\ref{fig:motivation}(b)). This indicates that the model has not truly captured the geometric essence of the \textit{open} affordance, but has merely memorized object-specific patterns.
Second, encoding the entire object into a coarse global feature leads to insufficient differentiation within its geometry. The lack of clear separation between object parts prevents the model from establishing clear semantic correspondence with the text (Fig.~\ref{fig:motivation}(c)), thereby reducing alignment efficiency and hindering fine-grained region prediction.

To address these limitations, we propose a two-stage framework that builds alignment-friendly textual representations and learns unified semantic--geometric features for open-vocabulary 3D affordance grounding. In the first stage, we recover the missing part-level semantics from raw text: large language models generate alignment-oriented descriptions that highlight the functional aspects of each affordance—including unseen ones—allowing semantically related affordances (e.g., \textit{demonstrate} and \textit{display}) to be linked to the same object part (Fig.~\ref{fig:motivation}(a)). An alignment loss further enforces a consistent affordance $\rightarrow$ part-level text $\rightarrow$ 3D region pathway to support open-vocabulary generalization. In the second stage, we learn fine-grained object-agnostic geometric representations through two complementary components: (1) Affordance Prototype Aggregation (APA) mechanism, which constructs a shared prototype space capturing cross-object geometric consistency for each affordance, preventing the model from memorizing object-specific patterns (Fig.~\ref{fig:motivation}(b)); and (2) Intra-Object Relational Modeling (IORM) module, which enhances geometric differentiation within each object, enabling clearer semantic correspondence with part-focused text and supporting fine-grained grounding (Fig.~\ref{fig:motivation}(c)). All these designs operate collaboratively within a unified framework.

Formally, given a natural language question and a 3D point cloud, our model predicts a per-point affordance probability mask. We first convert the question into a structured, part-aware instruction using a large language model, and encode it into token-level features. In parallel, geometric features are extracted from the point cloud and enhanced through the intra-object relational modeling module. The semantic and geometric features are then fused through cross-modal attention and refined via modulation mechanisms to predict the final mask. During training, we align part-level semantics with 3D structures and supervise functional abstraction through an affordance prototype aggregation mechanism.


To evaluate our method, we introduce a new large-scale dataset that offers comprehensive point cloud coverage, fine-grained mask annotations, and well-defined evaluation protocols. Extensive experiments on this dataset, together with two existing benchmarks, demonstrate the effectiveness of our approach.

Our main contributions are as follows:
\begin{itemize}
    \item We propose a cross-modal framework that learns unified semantic–geometric representations for open-vocabulary 3D affordance grounding, enabling robust functional generalization.
    \item We introduce two key modules—Affordance Prototype Aggregation(APA) and Intra-Object Relational Modeling (IORM)—to achieve cross-object functional consistency and enhance part-level geometric differentiation for precise semantic alignment.
    \item We present a new benchmark with large-scale point cloud coverage, fine-grained affordance annotations, and comprehensive protocol support. Extensive experiments demonstrating the effectiveness of our approach.
\end{itemize}

%% file: sec/2_related_works.tex
\section{Related works}
\label{sec:related}

\begin{figure*}[t]
    \centering
    \includegraphics[width=0.9\linewidth]{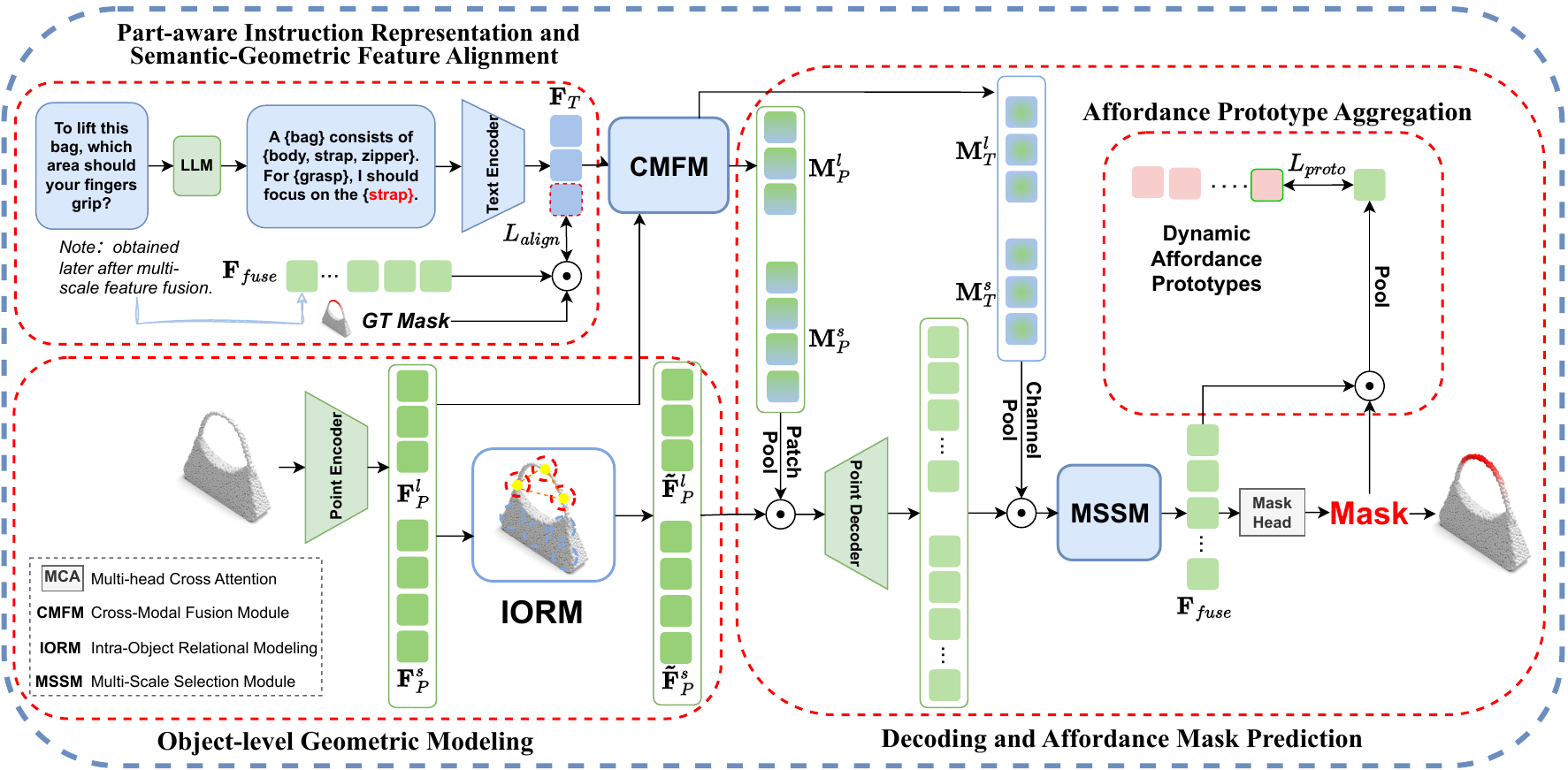}
    \caption{
Overview of our framework, comprising five stages: (1) Part-aware Instruction Representation (\S\ref{sec:PIGSGA}); (2) Object-level Geometric Modeling (\S\ref{sec:PEIORM}); (3) Cross-modal Fusion (\S\ref{sec:CMFM}); (4) Decoding and Mask Prediction (\S\ref{sec:mask}); and (5) Prototype Aggregation (\S\ref{sec:proto}). Zoom-in views of IORM, CMFM, and MSSM are in the Appendix.
}

    \label{fig:overview}
\end{figure*}

\subsection{3D Affordance Grounding}
Grounding functional regions in 3D objects from natural language has gained increasing attention~\cite{Li2024LASO, Yang2023IAG, Yu2025SeqAfford, Gao2025MIFAG, Zhu2025LMA}. Early methods treat affordance detection as point cloud classification, ignoring language~\cite{Deng2021AffordanceNet, Tabib2024LGAffordNet}. IAGNet~\cite{Yang2023IAG} projects 2D interaction images into 3D but relies on manual correspondences and closed vocabularies. LASO~\cite{Li2024LASO} introduces language-guided segmentation with a large dataset, but its fusion strategy and closed label set limit generalization. SeqAfford~\cite{Yu2025SeqAfford} extends grounding to sequential instructions, yet remains confined to a closed-set regime. MIFAG~\cite{Gao2025MIFAG} integrates affordance knowledge from multiple references but still depends on discrete labels.
Open-vocabulary affordance grounding methods have emerged. OpenAD~\cite{Nguyen2023OpenAD} aligns CLIP embeddings with point features for zero-shot detection, but is limited to binary masks and explicit tags. 3D-AffordanceLLM~\cite{Chu2025AffordanceLLM} reframes affordance detection as instruction reasoning with a language model and custom decoder, but does not capture fine-grained part–function alignment.

\subsection{Text-Point Cloud Cross-Modal Learning}
Cross-modal alignment between language and 3D point clouds underpins grounded 3D understanding. ScanRefer~\cite{Chen2020} and ReferIt3D~\cite{Achlioptas2020} define object localization from language queries. Extensions such as ScanQA~\cite{Azuma2022} and SQA3D~\cite{Ma2023} incorporate question answering for spatial and semantic reasoning. Dual-encoder models combine 3D features (e.g., PointNet++~\cite{qi2017pointnet++}) and pretrained language models (e.g., BERT~\cite{Devlin2019}), with fusion achieved via attention~\cite{Chen2022LCSR}, graph reasoning~\cite{Zhao2021}, or region refinement~\cite{luo20223DSPS}. Joint modeling of object–language relations enhances localization~\cite{Zhao2021, Jain2022}.
Recent work extends cross-modal learning to functional region segmentation. AffordanceNet~\cite{Deng2021AffordanceNet} and LASO~\cite{Li2024LASO} segment affordance regions from point clouds via instruction. LMAffordance3D~\cite{Zhu2025LMA} fuses visual observations and language for latent functionality inference. Large-scale models such as ULIP~\cite{Xue2023} and OpenShape~\cite{Liu2023} enable open-vocabulary capabilities through CLIP-style cross-modal alignment, supporting zero-shot 3D understanding.

\subsection{3D Open Vocabulary Learning}
Recent advances in 3D open-vocabulary learning enable zero-shot recognition and grounding of novel concepts via cross-modal supervision. OpenScene~\cite{Peng2023OpenScene} and PLA~\cite{Ding2023PLA} align 3D features with CLIP for zero-shot segmentation, while AIDE~\cite{Wang2025AIDE} and LangSplat~\cite{Qin2024LangSplat} further enhance language–geometry fusion through prompt optimization. For instance-level tasks, OpenMask3D~\cite{Takmaz2023OpenMask3D} and OpenIns3D~\cite{Huang2024OpenIns3D} match 3D proposals to CLIP queries. In detection,~\cite{Lu2023OV3DET, Yang2024ImOV3D} transfer 2D supervision, and methods like POP-3D~\cite{Vobecky2023POP3D} and LERF~\cite{Kerr2023LERF} embed language into volumetric or radiance field representations for free-form grounding. These works collectively unify 3D geometry and semantics beyond predefined label sets.

%% file: sec/3_method.tex
\section{Method}
\subsection{Overview}
\label{sec:overview}
We address the task of language-driven 3D affordance grounding. Formally, given a natural language question $x_{\text{lang}}$ and an object point cloud $P \in \mathbb{R}^{N \times 3}$ with $N$ points, the goal is to predict a per-point probability mask $\hat{\mathbf{M}} \in [0, 1]^N$ that localizes the functional region relevant to the question.
To address this task, we propose a semantics-guided framework that integrates structured instruction modeling, intra-object relational modeling, and affordance prototype aggregation to enable open-vocabulary, fine-grained 3D affordance grounding. To achieve this, our framework learns a unified semantic–geometric representation that binds part-aware language priors with structurally grounded 3D features, enabling functional generalization across both concepts and shapes. 

An overview of our architecture is shown in Fig. ~\ref{fig:overview}.
We first convert the question into a part-aware prompt using a large language model, and extract token-level features via a text encoder (Sec.~\ref{sec:PIGSGA}). In parallel, multi-scale geometric features are extracted from the point cloud using a point backbone and enhanced through an intra-object relational modeling module (Sec.~\ref{sec:PEIORM}).
These features are fused via a dual-stage cross-modal fusion module, enabling mutual semantic–geometric interaction (Sec.~\ref{sec:CMFM}). Original point features are decoded and modulated by the fused features to produce the final point-wise representations, from which an MLP predicts the affordance mask (Sec.~\ref{sec:mask}).
Finally, to enable abstraction over structurally diverse but functionally similar parts, the predicted region is embedded and aligned with a learnable prototype set representing canonical affordance concepts (Sec.~\ref{sec:proto}). 

\subsection{Part-aware Instruction Representation}
\label{sec:PIGSGA}

Given a raw natural language question (e.g., \textit{``To lift this bag, which area should your fingers grip?''}), we use a large language model to generate a structured, part-aware instruction (e.g., \textit{``A bag consists of body, strap, zipper. For grasp, I should focus on the strap.''}), which clarifies the intent and the relevant object part.
The structured instruction $x_{\text{struct}}$ is encoded by a pre-trained RoBERTa model~\cite{liu2019roberta}:
\begin{equation}
\mathbf{F}_{\text{T}} = f_{\text{text}}(x_{\text{struct}}) \in \mathbb{R}^{C \times L},
\end{equation}
where part-specific token (e.g., ``strap'') and its feature embedding $T_i$ are extracted to enable part-aware semantic–geometric alignment during training.

\subsection{Object-level Geometric Modeling}
\label{sec:PEIORM}

To capture both global structures and local details, we extract multi-scale geometric features from the input point cloud using a hierarchical PointNet++ encoder~\cite{qi2017pointnet++}. Inspired by \cite{obj_afford}, the input point cloud $P$ is downsampled into large-scale and small-scale region sets, $P^l$ and $P^s$, producing multi-scale features $\mathbf{F}_P^l \in \mathbb{R}^{C \times N_P^l}$ and $\mathbf{F}_P^s \in \mathbb{R}^{C \times N_P^s}$, where $C$ is the feature dimension and $N_P^l$, $N_P^s$ are the number of regions at the large and small scales, respectively.

To further reinforce connectivity and feature consistency among regions of an object, we introduce a intra-object relational modeling module (IORM). For each region $i$, we first compute scaled dot-product similarities between its query vector $\mathbf{q}_i$ and the key vectors $\mathbf{k}_j$ of all other regions:
\begin{equation}
\mathbf{S}_{ij} = \frac{\mathbf{q}_i^\top \mathbf{k}_j}{\sqrt{C}}.
\end{equation}
We then select the top-$k$ most similar regions $\mathcal{N}_k(i)$ based on $\mathbf{S}_{ij}$, and aggregate their corresponding value features $\mathbf{v}_j$ to update the representation of region $i$:
\begin{equation}
\tilde{\mathbf{f}}_i = \sum_{j \in \mathcal{N}_k(i)} \text{softmax}_j(\mathbf{S}_{ij}) \cdot \mathbf{v}_j.
\end{equation}
This process is applied at both scales, yielding the enhanced features $\tilde{\mathbf{F}}_P^l\in \mathbb{R}^{C \times N_P^l}$ and $\tilde{\mathbf{F}}_P^s\in \mathbb{R}^{C \times N_P^s}$.

\subsection{Cross-modal Fusion Module}
\label{sec:CMFM}

To bridge linguistic semantics and geometric structure, we propose a dual-stage cross-modal fusion module (CMFM) that performs bidirectional interaction between textual and geometric features across both scales. CMFM comprises two cascaded multi-head cross-attention blocks: (1) \textit{text-to-point fusion}, which injects affordance semantics into geometric features, and (2) \textit{point-to-text fusion}, which incorporates structural cues into the semantic stream.
\paragraph{Stage 1: Text-to-Point Fusion.}
Given the structured text embedding $\mathbf{F}_{\text{T}} \in \mathbb{R}^{C \times L}$ from Section~\ref{sec:PIGSGA} and the multi-scale point features $\mathbf{F}_P^{l/s} \in \mathbb{R}^{C \times N_P^{l/s}}$ from Section~\ref{sec:PEIORM}, we treat point features as queries and text embeddings as keys and values. Cross-attention is applied at each scale:
\begin{equation}
\mathbf{M}_P^{l/s} = \text{MHAttn}(\mathbf{F}_P^{l/s}, \mathbf{F}_{\text{T}}, \mathbf{F}_{\text{T}}).
\end{equation}
This injects affordance semantics into geometric features.

\paragraph{Stage 2: Point-to-Text Fusion.}
We then treat $\mathbf{M}_P^{l/s}$ as keys and values, and $\mathbf{F}_{\text{T}}$ as queries:
\begin{equation}
\mathbf{M}_T^{l/s} = \text{MHAttn}(\mathbf{F}_{\text{T}}, \mathbf{M}_P^{l/s}, \mathbf{M}_P^{l/s}).
\end{equation}
The resulting $\mathbf{M}_T^{l/s}$ capture structure-aware affordance semantics for downstream grounding. Each path employs standard multi-head attention.

\subsection{Decoding and Affordance Mask Prediction}
\label{sec:mask}
After cross-modal fusion and intra-relational modeling, we obtain enhanced point features $\tilde{\mathbf{F}}_P^{l/s}$, semantic-injected geometric features $\mathbf{M}_P^{l/s}$ and structure-aware semantic features $\mathbf{M}_T^{l/s}$ . We first perform patch-wise modulation by applying element-wise multiplication between $\tilde{\mathbf{F}}_P^{l/s}$ and patch-aggregated semantic derived from $\mathbf{M}_P^{l/s}$. The resulting features are then upsampled via PointNet++ Feature Propagation (FP)~\cite{qi2017pointnet++}. Finally, channel-wise modulation is applied using semantic cues aggregated from $\mathbf{M}_T^{l/s}$:
\begin{equation}
\hat{\mathbf{G}}^{l/s} = \text{FP}\left(\tilde{\mathbf{F}}_P^{l/s} \odot \text{Pool}_{\text{patch}}(\mathbf{M}_P^{l/s})\right) \odot \text{Pool}_{\text{chan}}(\mathbf{M}_T^{l/s}).
\end{equation}
To  balance coarse and fine geometric cues, we employ a Multi-Scale Selection Module (MSSM), which computes soft weights $\alpha_l, \alpha_s$ ($\alpha_l + \alpha_s = 1$) for gated fusion:
\begin{equation}
\mathbf{F}_{\text{fuse}} = \alpha_l \hat{\mathbf{G}}^l + \alpha_s \hat{\mathbf{G}}^s.
\end{equation}
A MLP mask head $f_{\text{mask}}$ produces the per-point affordance probability mask:
\begin{equation}
\hat{\mathbf{M}} = \sigma\left(f_{\text{mask}}(\mathbf{F}_{\text{fuse}})\right)\in \mathbb{R}^{N \times 1},
\end{equation}
where $\sigma(\cdot)$ denotes the sigmoid activation.

\subsection{Affordance Prototype Aggregation}
\label{sec:proto}

To embed functionally consistent regions into a shared prototype space, we introduce an Affordance Prototype Aggregation (APA) mechanism. Given the fused point-wise features $\mathbf{F}_{\text{fuse}} \in \mathbb{R}^{C \times N}$ and predicted mask scores $\hat{\mathbf{M}}$ (both from Section~\ref{sec:mask}), we compute a region embedding $\mathbf{z} \in \mathbb{R}^C$ via masked average pooling over $\mathbf{F}_{\text{fuse}}$, where each point's contribution is weighted by its corresponding score in $\hat{\mathbf{M}}$.
A learnable prototype set $\mathbf{P} = [\mathbf{p}_1, \dots, \mathbf{p}_K]^\top \in \mathbb{R}^{K \times C}$, randomly initialized at the start of training, encodes canonical affordance concepts and is dynamically extended during training as novel affordances emerge, eliminating the need to predefine the number of prototypes. For each region embedding $\mathbf{z}$, its similarity to each prototype is computed via cosine similarity:
\begin{equation}
s_k = \frac{\mathbf{z}^\top \mathbf{p}_k}{\|\mathbf{z}\|\,\|\mathbf{p}_k\|}.
\end{equation}
We optimize the association between region embeddings and prototypes during training to enable effective affordance prototype aggregation.

\begin{table*}[t]
\small
\centering

\caption{Public 3D affordance datasets in recent literature.
\textsuperscript{\dag}\,Only a total of 180K points–text pairs is reported; instance‑level counts are not specified. %
\textsuperscript{\ddag}\,Dataset provides 42119 points–text pairs without separate point cloud instance statistics.}
\label{tab:dataset-overview}

\begin{tabular}{lccccc}
\toprule
\textbf{Dataset} & \textbf{\#Point Cloud Inst.} & \textbf{\#Question Inst.}
                 & \textbf{\#Image Inst.} & \textbf{OV‑Affordance} & \textbf{Probabilistic Mask} \\
\midrule
3DAffordanceNet                          & 23K  & \xmark  & \xmark & \xmark & \cmark \\
PIAD                        & 7K  & \xmark  & 5162& \xmark & \cmark \\
LASO                          & 8.4K  & 870    & \xmark & \xmark & \cmark \\
Seq‑AFD                      & 18K & \dash \textsuperscript{\dag} & \xmark & \xmark & \cmark \\

\rowcolor{gray!10} Ours                 & 23K & 1840& \xmark & \cmark & \cmark \\
\bottomrule
\end{tabular}
\end{table*}

%% file: sec/4_training.tex
\section{Training and Inference}
\label{sec:training}
\subsection{Training Objective}
We train our model in an end-to-end manner using three loss objectives, which are detailed below.
\paragraph{Part-aware Semantic–Geometry Alignment Loss.}
To enforce alignment between part-level semantics and 3D geometry, we compute a cosine similarity loss between the target part embedding $T_i$ (Section~\ref{sec:PIGSGA}) and the ground-truth region embedding $\mathbf{G}_{\text{gt}}$. We obtain $\mathbf{G}_{\text{gt}}$ via masked average pooling over fused features $\mathbf{F}_{\text{fuse}}$ (Section~\ref{sec:mask}), weighted by the ground-truth mask $\mathbf{M}_{\text{gt}}$:
\begin{equation}
\mathcal{L}_{\text{align}} = 1 - \frac{T_i^\top \mathbf{G}_{\text{gt}}}{\|T_i\| \, \|\mathbf{G}_{\text{gt}}\|}.
\label{eq:align}
\end{equation}
\paragraph{Prototype Association Loss.}
We guide region embeddings toward affordance prototypes by optimizing their association with a learnable prototype set via temperature-scaled cross-entropy:
\begin{equation}
\mathcal{L}_{\text{proto}} = -\log\frac{\exp(s_{y_{\text{aff}}}/\tau)}{\sum_{k=1}^{K}\exp(s_k/\tau)},
\label{eq:proto}
\end{equation}
We compute $s_k$ as described in Section~\ref{sec:proto}, $\tau$ is a temperature parameter, and $y_{\text{aff}}$ denotes the target affordance class derived from the structured instruction.
\paragraph{Affordance Mask Loss.}
For affordance mask prediction, a hybrid loss is computed between the predicted affordance mask $\hat{\mathbf{M}}$ and ground-truth mask $\mathbf{M}_{\text{gt}}$:
\begin{equation}
\mathcal{L}_{\text{mask}} = \mathcal{L}_{\text{focal}}(\hat{\mathbf{M}}, \mathbf{M}_{\text{gt}}) +  \mathcal{L}_{\text{dice}}(\hat{\mathbf{M}}, \mathbf{M}_{\text{gt}}),
\label{eq:mask}
\end{equation}
where focal loss~\cite{lin2017focal} addresses class imbalance and dice loss~\cite{milletari2016vnet} encourages region-level alignment. Detailed loss formulations are provided in Appendix.

\paragraph{Total Loss.}
The overall loss combines all the objectives:
\begin{equation}
\mathcal{L}_{\text{total}} = \mathcal{L}_{\text{mask}} + \beta_1 \mathcal{L}_{\text{align}} + \beta_2 \mathcal{L}_{\text{proto}},
\label{eq:total}
\end{equation}
where $\beta_1$ and $\beta_2$ are balancing hyperparameters, and $\mathcal{L}_{\text{mask}}$  $\mathcal{L}_{\text{align}}$ , and $\mathcal{L}_{\text{proto}}$ are defined above in Eq.~\ref{eq:mask},\ref{eq:align} and \ref{eq:proto}. 
\subsection{Inference}
At inference time, we remove all loss-related branches and auxiliary supervision. The model predicts the per-point affordance mask $\hat{\mathbf{M}}$ given an input question and point cloud, leveraging the unified semantic–geometric representations learned during training. Prototype association and semantic–geometry alignment are used only as training regularizers and do not affect inference efficiency.

%% file: sec/5_experiments.tex
\section{Experiments}
\label{sec:experiment}
\subsection{Datasets}
\label{sec:datasets}
\subsubsection{Proposed Dataset}
\label{sec:proposeddataset}
To overcome limitations of existing datasets and enable a comprehensive evaluation of our method, we introduce OpenAfford, a language-driven 3D affordance grounding benchmark, comprising over 23k point clouds with probabilistic affordance masks and 1,840 diverse natural language questions. As shown in Table~\ref{tab:dataset-overview}, OpenAfford uniquely supports open-vocabulary grounding with fine-grained probabilistic masks and comprehensive evaluation protocols. It provides significantly more point cloud annotations than LASO~\cite{Li2024LASO} and PIAD~\cite{Yang2023IAG}, supports open-vocabulary evaluation unlike LASO and Seq-AFD~\cite{Yu2025SeqAfford}, and improves upon the binary-labeled 3D-AffordanceLLM~\cite{Chu2025AffordanceLLM} by enabling generalization measurement to unseen object categories—crucial for assessing performance in open-world scenarios.
Point cloud labels are adapted from 3D AffordanceNet~\cite{Deng2021AffordanceNet}, and questions are drawn from LASO for closed-set and GPT-4o-generated queries for open-vocabulary testing (generation details are provided in the Appendix). To comprehensively assess generalization, we define two main evaluation settings, each with two subsets: (1) \textit{Closed-set evaluation}, including (a) \textit{Seen}, where both object categories and affordance types appear during training, and (b) \textit{Unseen}, where the test set includes six object categories not seen during training, measuring generalization to novel shapes under familiar affordance types; (2) \textit{Open-set evaluation}, including (a) \textit{Full-view}, where the test set includes sixteen affordance types unseen during training with full geometry available for zero-shot grounding, and (b) \textit{Partial-view}, where only partial object observations are provided, simulating incomplete scans and further challenging model robustness. Detailed data splits, including point cloud–affordance pairs, and the distribution of unseen affordance types and object categories across four evaluation settings, are provided in the Appendix.
These settings systematically probe semantic, categorical, and geometric generalization, establishing a unified benchmark for open-world 3D affordance understanding.

\subsubsection{External Datasets}
For cross-dataset validation, we further evaluate our model on two external benchmarks: the LASO dataset~\cite{Li2024LASO} under its Seen split, and the 3D-AffordanceLLM benchmark~\cite{Chu2025AffordanceLLM}. The LASO benchmark is limited to closed-set affordance categories, providing a controlled setting for transfer evaluation, whereas 3D-AFDLLM supports open-vocabulary affordance queries, enabling assessment of generalization to unseen affordances.


\subsection{Experimental Setting}
\subsubsection{Baselines and Evaluation Metrics}
\label{sec:baselines}
On the OpenAfford benchmark, we compare our method with recent 3D affordance grounding approaches, including LASO~\cite{Li2024LASO}, OpenAD~\cite{Nguyen2023OpenAD},IAGNet~\cite{Yang2023IAG} and a strong 3D understanding baseline XMF~\cite{aiello2022XMF}. All methods are evaluated under a unified protocol with the same backbone; model adaptations are detailed in the Appendix.
On the LASO benchmark, we compared against results from the original paper~\cite{Li2024LASO}, including ReferTrans~\cite{Li_Sigal_2021ReferTrans}, ReLA~\cite{liu2023ReLA}, 3D-SPS~\cite{luo20223DSPS} and LASO~\cite{Li2024LASO}. 
And on the 3D-AffordanceLLM benchmark, we compared against results from the original paper~\cite{Chu2025AffordanceLLM}, including ShapeLLM~\cite{qi12shapellm}, OpenAD~\cite{Nguyen2023OpenAD}, IAGNet~\cite{Yang2023IAG}, LASO~\cite{Li2024LASO} and 3D-AffordanceLLM~\cite{Chu2025AffordanceLLM}.

Following prior work~\cite{Yang2023IAG, Gao2025MIFAG, Li2024LASO}, we report AUC~\cite{Fawcett2006ROC}, aIoU, SIM~\cite{Fan2017StructureMeasure} and MAE on OpenAfford and LASO benchmark, where higher is better except for MAE.
For the 3D-AffordanceLLM benchmark, we follow the official evaluation protocol and report mIoU (mean IoU over all classes), Acc (overall point-wise accuracy), and mAcc (mean class-wise accuracy).
Metric definitions are shown in Appendix.
\subsubsection{Implementation Details}
All models are trained on a single NVIDIA H20 GPU. We use batch sizes of 96/32/8 and learning rates of 6e-4/1e-4/3e-4 for OpenAfford, LASO and IRAS, respectively. We optimize all parameters with the Adam optimizer~\cite{kingma2015adam}. Full training details are in the Appendix.

\begin{figure*}[t]
\vspace{-1em}
    \centering
    \includegraphics[width=0.95\linewidth]{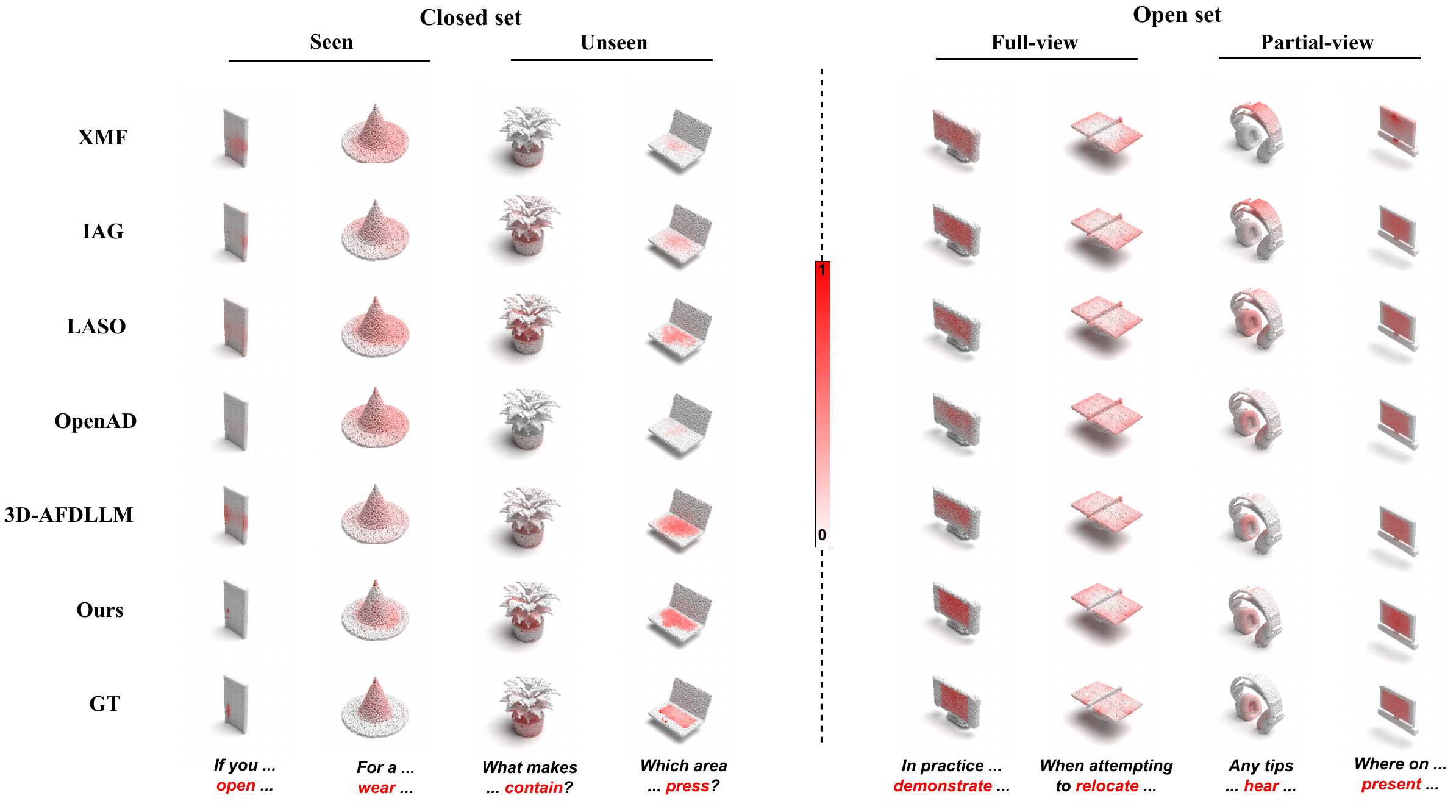}
    \caption{Visualization comparison. Results under four evaluation splits with predictions from each method. Red: predicted regions.}
    \label{fig:vismain}
    \vspace{-1em}
\end{figure*}

\begin{table}[t]
\centering
\small
\caption{Quantitative comparison across four evaluation splits using aIoU, AUC, SIM, and MAE. Best results are in bold.}
\label{tab:comparison}
\begin{tabular}{cc l cccc}
\toprule
\multicolumn{2}{c}{} & \textbf{Method} & \textbf{aIoU~$\uparrow$} & \textbf{AUC~$\uparrow$} & \textbf{SIM~$\uparrow$} & \textbf{MAE$\downarrow$} \\
\midrule
\multirow{10}{*}{\rotatebox{90}{\textbf{Open-set}}} 
  & \multirow{5}{*}{\rotatebox{90}{Full-view}} 
    & XMF         &      11.37&      76.33&      0.529&      0.117\\
  & & IAG         &      9.63&      74.95&      0.505&      0.123\\
  & & LASO        &      13.86&      76.12&      0.540&      \textbf{0.109}\\
  & & OpenAD      &      12.57&      79.80&      0.570&      0.113\\
  & & \cellcolor{gray!15}Ours&      \cellcolor{gray!15}\textbf{18.38}&      \cellcolor{gray!15}\textbf{82.33}&      \cellcolor{gray!15}\textbf{0.606}&      \cellcolor{gray!15}0.110\\
\cmidrule(lr){2-7}
  & \multirow{5}{*}{\rotatebox{90}{Partial-view}} 
    & XMF         &      9.03&      67.54&      0.450&      0.127\\
  & & IAG         &      8.11&      68.31&      0.433&      0.122\\
  & & LASO        &      9.70&      69.11&      0.477&      0.129\\
  & & OpenAD      &      11.49&      79.16&      0.559&      0.104\\
  & & \cellcolor{gray!15}Ours &\cellcolor{gray!15}\textbf{15.85}&      \cellcolor{gray!15}\textbf{83.17}&      \cellcolor{gray!15}\textbf{0.578}&      \cellcolor{gray!15}\textbf{0.101}\\

\midrule
\multirow{10}{*}{\rotatebox{90}{\textbf{Closed-set}}} 
  & \multirow{5}{*}{\rotatebox{90}{Seen}} 
    & XMF         &      12.76&      81.49&      0.553&      0.114
\\
  & & IAG         &      16.73&      83.87&      0.593&      0.100
\\
  & & LASO        &      17.66&      85.43&      0.613&      0.097
\\
  & & OpenAD      &      16.89&      85.89&      0.618&      0.095\\
  & & \cellcolor{gray!15}Ours&      \cellcolor{gray!15}\textbf{19.18}&      \cellcolor{gray!15}\textbf{86.69}&      \cellcolor{gray!15}\textbf{0.628}&      \cellcolor{gray!15}0.106\\
\cmidrule(lr){2-7}
  & \multirow{5}{*}{\rotatebox{90}{Unseen}} 
    & XMF         &      12.29&      80.92&      0.552&      0.116\\
  & & IAG         &      14.15&      81.28&      0.559&      0.117\\
  & & LASO        &      17.37&      85.14&      0.602&      \textbf{0.099}\\
  & & OpenAD      &      16.88&      83.61&      0.605&      0.102\\
  & & \cellcolor{gray!15}Ours&      \cellcolor{gray!15}\textbf{17.81}&      \cellcolor{gray!15}\textbf{85.22}&      \cellcolor{gray!15}\textbf{0.608}&      \cellcolor{gray!15}0.100\\
\bottomrule
\end{tabular}
\end{table}

\subsection{Comparison Results}
\label{sec:quant_results}

\subsubsection{Main Results on OpenAfford}
\label{sec:comparison}
\paragraph{Quantitative Analysis}
We evaluate our method against recent approaches on the OpenAfford benchmark under four settings, providing a comprehensive assessment of model performance across diverse scenarios. As shown in Table~\ref{tab:comparison}, our model consistently outperforms all baselines in aIoU, AUC, and SIM, while maintaining low MAE across splits.
In the Open-set Full-view setting, where models must infer unseen affordances, our approach achieves the highest aIoU (18.38, +32.6\% over the best baseline), AUC (82.33), and SIM (0.606). Such a substantial gain demonstrates that our method effectively transfers affordance understanding beyond the training vocabulary. The improvement mainly stems from the structual semantic–geometric alignment and affordance prototype aggregation, which enable our model to generalize to novel functional cues even when explicit supervision is unavailable. Similar gains are observed in the Partial-view setting, where the input geometry is incomplete; our model improves aIoU by +37.9\%, highlighting ability to infer missing structural information and maintain coherent region-level reasoning under partial observations.
In Closed-set settings, our model also leads on both Seen and Unseen splits, achieving +13.6\% and +2.5\% relative improvements in aIoU, respectively. This shows that our framework not only captures the functional semantics of seen objects, but also generalizes effectively to unseen object geometries.
These results validate the effectiveness of our structural semantic alignment, intra-object relational modeling, and affordance-guided prototype aggregation in achieving precise and generalizable affordance grounding. Hyperparameter sensitivity analysis is provided in the Appendix.

\vspace{-1em}
\paragraph{Qualitative Analysis}
We visualize representative predictions across all four settings in Figure~\ref{fig:vismain}. Our model produces precise masks that align well with the language intent.
In Open-set cases like “demonstrate”, the model grounds unseen affordances to appropriate regions. Under Partial-view inputs, predictions remain robust despite missing geometry, showing strong structural generalization.
Compared to other methods, our results better match the ground truth in terms of boundary completeness and region consistency, validating both semantic grounding and geometric reasoning. More randomly selected visualizations are provided in Appendix.

\subsubsection{Cross-Datasets Evaluation}
\label{sec:cross_laso}
We train our model from scratch on LASO benchmark~\cite{Li2024LASO} under its Seen split. As shown in Table~\ref{tab:benchmark-laso}, our model matches LASO’s aIoU and exceeds all the competitors in AUC, SIM, and MAE
We also train our model from scratch on 3D-AffordanceLLM benchmark~\cite{Chu2025AffordanceLLM}. As shown in Table~\ref{tab:3D-AFDLLM}, our approach achieves the best performance in both full-view and partial-view settings.
The results on both datasets demonstrate strong adaptability to new domains.The qualitative analysis is shown in Appendix.

\begin{table}[t]
\centering
\small
\caption{Quantitative results on the LASO dataset (Seen split). Results for all baselines are reported from the official LASO paper~\cite{Li2024LASO}. }
\label{tab:benchmark-laso}
\begin{tabular}{lcccc}
\toprule
\textbf{Method} & \textbf{aIoU$\uparrow$} & \textbf{AUC$\uparrow$} & \textbf{SIM$\uparrow$} & \textbf{MAE$\downarrow$} \\
\midrule
LASO        & 20.8 & 87.3 & 0.629 & 0.093 \\
IAG         & 17.8 & 82.3 & 0.561 & 0.109 \\
3D-SPS      & 11.4 & 76.2 & 0.433 & 0.138 \\
ReLA        & 15.2 & 78.9 & 0.532 & 0.118 \\
ReferTrans  & 13.7 & 79.8 & 0.497 & 0.124 \\
\rowcolor{gray!15}
Ours & \textbf{20.8} & \textbf{88.8} & \textbf{0.637} & \textbf{0.090} \\
\bottomrule
\end{tabular}
\end{table}

\begin{table}[t]
\centering
\small

\caption{Quantitative results on the 3D-AffordanceLLM dataset. Results for all baselines are reported from the official paper~\cite{Chu2025AffordanceLLM}.}
\label{tab:3D-AFDLLM}
\begin{tabular}{c l ccc}
\toprule
\multicolumn{1}{c}{} & \textbf{Method} & \textbf{mIoU~$\uparrow$} & \textbf{Acc~$\uparrow$} & \textbf{mAcc~$\uparrow$} \\
\midrule
\multirow{8}{*}{\rotatebox{90}{Full-view}} 
  & ShapeLLM & 0.88 & 0.28 & 0.99 \\
  & OpenAD-PointNet++ & 13.53 & 3.97 & 16.40 \\
  & OpenAD-DGCNN & 11.15 & 3.84 & 13.86 \\
  & IAGNet & 16.16 & 19.07 & 23.92 \\
  & LASO & 22.41 & 15.90 & 30.22 \\
  & 3D-ADLLM-Qwen & 24.43 & 23.90 & 35.45 \\
  & 3D-ADLLM-Phi & 30.43 & 29.36 & \textbf{47.78} \\
  & \cellcolor{gray!15}Ours & \cellcolor{gray!15}\textbf{32.15} & \cellcolor{gray!15}\textbf{31.18} & \cellcolor{gray!15}45.97 \\
\midrule
\multirow{8}{*}{\rotatebox{90}{Partial-view}} 
  & ShapeLLM & 1.49 & 1.35 & 1.70 \\
  & OpenAD-PointNet++ & 11.29 & 2.41 & 13.88 \\
  & OpenAD-DGCNN & 8.04 & 1.58 & 9.85 \\
  & IAGNet & 14.36 & 16.90 & 21.73 \\
  & LASO & 20.06 & 8.80 & 26.84 \\
  & 3D-AFDLLM-Qwen & 26.25 & 29.50 & 41.57 \\
  & 3D-AFDLLM-Phi & 27.25 & 27.87 & 39.04 \\
  & \cellcolor{gray!15}Ours & \cellcolor{gray!15}\textbf{30.22} & \cellcolor{gray!15}\textbf{29.15} & \cellcolor{gray!15}\textbf{41.66} \\
\bottomrule
\end{tabular}
\end{table}

\begin{table}[t]
\centering
\small
\caption{Ablation results under all four evaluation splits. We compare the full model with variants missing key components. Performance is measured using aIoU, AUC, SIM, and MAE.}
\label{tab:ablation}
\begin{tabular}{cc l cccc}
\toprule
\multicolumn{2}{c}{} & \textbf{Method} & \textbf{aIoU$\uparrow$} & \textbf{AUC$\uparrow$} & \textbf{SIM$\uparrow$} & \textbf{MAE$\downarrow$} \\
\midrule
\multirow{10}{*}{\rotatebox{90}{\textbf{Open-set}}} 
  & \multirow{5}{*}{\rotatebox{90}{Full-view}} 
    & w/o PIG&      12.67&      80.09&      0.545&      0.108\\
  & & w/o APA&      14.31&      82.30&      0.585&      0.105\\
  & & w/o PSGA&      14.16&      81.22&      0.568&      0.103\\
  & & w/o IORM&      17.36&      82.11&      0.599&      \textbf{0.102}\\
  & & \cellcolor{gray!15}Ours&      \cellcolor{gray!15}\textbf{18.38}&      \cellcolor{gray!15}\textbf{82.33}&      \cellcolor{gray!15}\textbf{0.606}&      \cellcolor{gray!15}0.110\\
\cmidrule(lr){2-7}
  & \multirow{5}{*}{\rotatebox{90}{Partial-view}} 
    & w/o PIG&      11.34&      75.99&      0.532&      0.113\\
  & & w/o APA&      14.76&      78.28&      0.565&      0.106\\
  & & w/o PSGA&      14.26&      80.28&      0.573&      0.101\\
  & & w/o IORM&      13.97&      78.56&      0.562&      0.108\\
  & & \cellcolor{gray!15}Ours &\cellcolor{gray!15}\textbf{15.85}&      \cellcolor{gray!15}\textbf{83.17}&      \cellcolor{gray!15}\textbf{0.578}&      \cellcolor{gray!15}\textbf{0.101}\\

\midrule
\multirow{10}{*}{\rotatebox{90}{\textbf{Closed-set}}} 
  & \multirow{5}{*}{\rotatebox{90}{Seen}} 
    & w/o PIG&      16.44&      84.92&      0.590&      0.104\\
  & & w/o APA&      17.37&      86.45&      0.614&      0.113\\
  & & w/o PSGA&      16.95&      86.63&      0.612&      \textbf{0.097}\\
  & & w/o IORM&      17.72&      86.50&      0.620&      0.099\\
  & & \cellcolor{gray!15}Ours&      \cellcolor{gray!15}\textbf{19.18}&      \cellcolor{gray!15}\textbf{86.69}&      \cellcolor{gray!15}\textbf{0.628}&      \cellcolor{gray!15}0.106\\
\cmidrule(lr){2-7}
  & \multirow{5}{*}{\rotatebox{90}{Unseen}} 
    & w/o PIG&      15.04&      83.51&      0.589&      0.103\\
  & & w/o APA&      16.05&      85.20&      0.598&      0.104\\
  & & w/o PSGA&      15.41&      85.07&      0.589&      0.106\\
  & & w/o IORM&      16.42&      85.16&      0.601&      0.102\\
  & & \cellcolor{gray!15}Ours&      \cellcolor{gray!15}\textbf{17.81}&      \cellcolor{gray!15}\textbf{85.22}&      \cellcolor{gray!15}\textbf{0.608}&      \cellcolor{gray!15}\textbf{0.100}\\
\bottomrule
\end{tabular}
\end{table}

\subsection{Ablation Study}
\label{sec:ablation}


We perform ablation experiments by removing components from the full model; implementation details are provided in the Appendix. The modules correspond to key design choices: PIG (Part-aware Instruction Generation), PSGA (Part-aware Semantic–Geometric Alignment), IORM (Intra-Object Relational Modeling), and APA (Affordance Prototype Aggregation). Results across all evaluation splits are shown in Table~\ref{tab:ablation}.
Removing PIG—on which PSGA and APA depend—leads to the largest drop in aIoU in the Open-set Full-view setting (–5.71, –31.1\%), highlighting the importance of structured language grounding. Disabling PSGA also causes a substantial degradation, especially in open-set scenarios, where aIoU decreases from 18.38 to 14.16 (–4.22, –22.9\%), indicating weakened generalization and reduced semantic focus.
Removing APA results in large declines across all four splits, with particularly severe drops in open-set settings. In the Full-view split, aIoU decreases by 3.87 (–21.1\%), and in the Partial-view split by 1.09 (–6.9\%), indicating a weakened ability to capture geometric consistency of the same affordance across different objects.
IORM primarily affects region completeness and prediction consistency: in the Partial-view setting, aIoU drops by 1.88 (–11.9\%) and MAE increases from 0.101 to 0.108 without IORM, indicating reduced capacity to model spatial relations among parts.
These results confirm the effectiveness of each module under four settings.

\begin{figure}[t]
    \centering
    \includegraphics[width=0.85\linewidth]{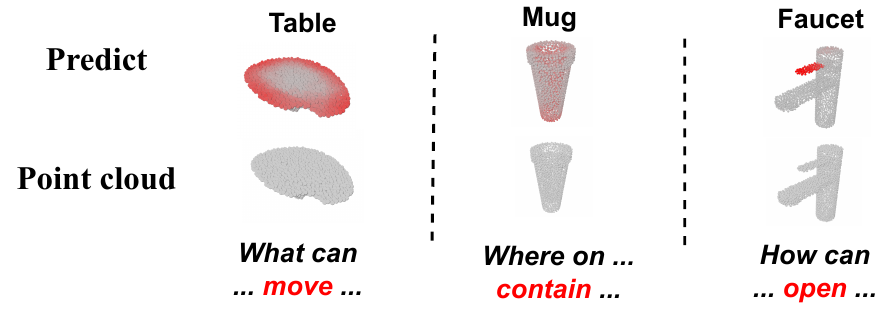}
  \caption{
Cross-dataset zero-shot predictions showing semantically consistent results across domain shifts.
}
    \label{fig:viscrossdataset}
    \vspace{-1em}
\end{figure}

\subsection{Cross-dataset Zero-shot Visualization}
\label{sec:cross_zero_visual}

We further evaluate our model on ShapeNet~\cite{chang2015shapenet} objects. As shown in Figure~\ref{fig:viscrossdataset}, our predictions remain semantically consistent despite the domain gap in geometry and rendering style. This demonstrates the model’s ability to generalize to new distributions in a zero-shot manner. Additional qualitative results are shown in Appendix.

%% file: sec/6_conclusion.tex
\section{Conclusion}

In this work, we propose a novel framework for open-vocabulary 3D affordance grounding, addressing open-vocabulary understanding, semantic–geometric alignment, and generalization. By leveraging large language models for part-aware instructions and introducing the Affordance Prototype Aggregation (APA) and Intra-Object Relational Modeling (IORM) modules, our method improves functional generalization and fine-grained region prediction. Experimental results on a new benchmark dataset, along with two existing benchmarks, demonstrate the superior performance of our approach.
\vspace{-4mm}
\paragraph{Limitations}
A primary limitation of our approach is its reliance on LLMs for question transformation, which can introduce latency and occasional misinterpretations. We analyze these issues in the Appendix and plan to address them by exploring fine-tuned lightweight models to improve efficiency and robustness.

%% file: sec/X_suppl.tex
\clearpage
\setcounter{page}{1}
\setcounter{section}{0}
\setcounter{figure}{0}
\setcounter{table}{0}
\renewcommand{\thefigure}{S\arabic{figure}}
\renewcommand{\thetable}{S\arabic{table}}
\maketitlesupplementary


\renewcommand{\thefigure}{S\arabic{figure}}
\renewcommand{\thetable}{S\arabic{table}}

\section*{Overview of Appendices}


In this supplementary material, we present additional implementation details and experimental results. First, we provide a more detailed description of the model in Sec. \ref{modeldetails}. Next, we include the implementation details of baseline methods, evaluation metrics, and training procedures, as well as additional ablation study details in Sec. \ref{baseline} - \ref{ablation}. Furthermore, in Sec. \ref{experiments}, we present further experiments, including hyperparameter sensitivity analysis, error analysis, and additional visualizations of the model's performance across different datasets.


\section{Model Details}
\label{modeldetails}
\subsection{More Architectural Details}

\begin{figure*}[t]
    \centering
    \includegraphics[width=1\linewidth]{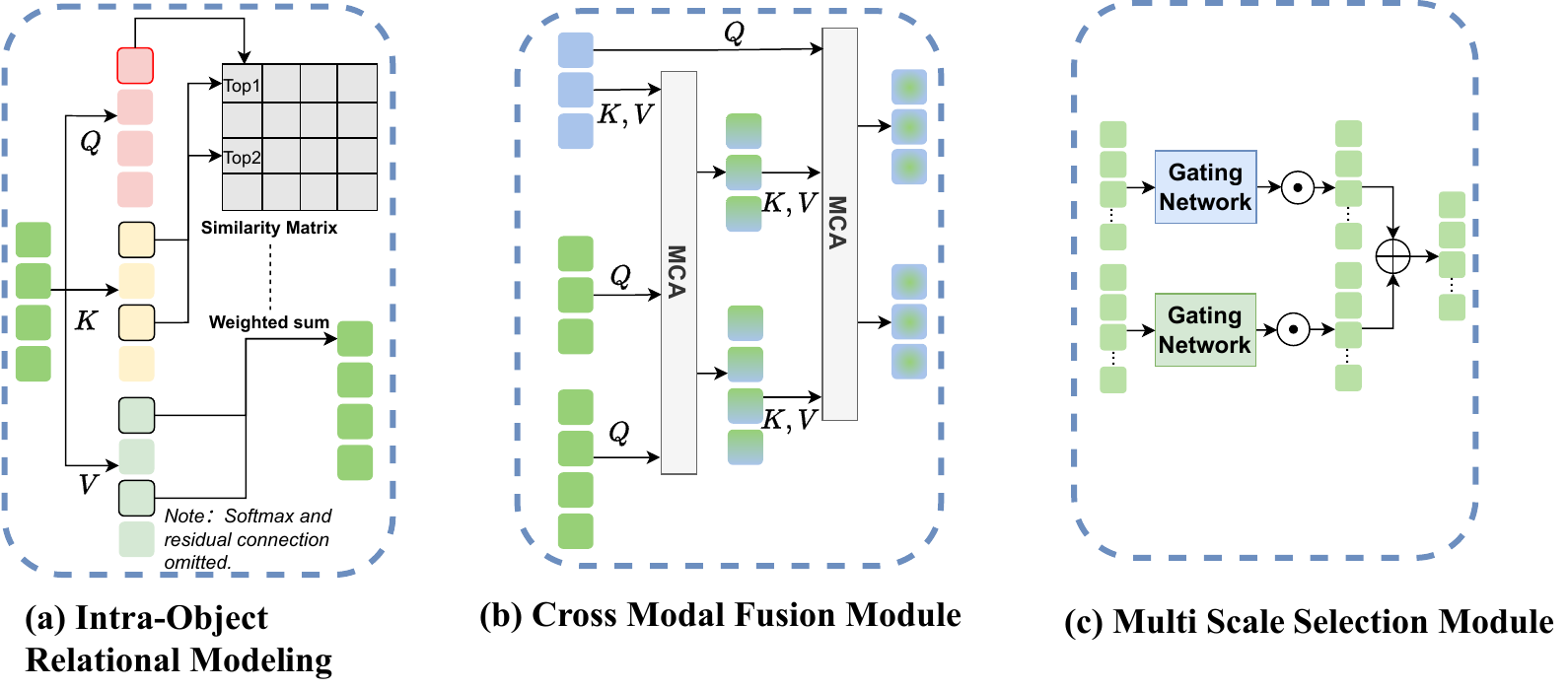}
    \caption{
        Model architecture details of three components: (a) Intra-Object Relational Modeling (IORM); (b) Cross-Modal Fusion Module (CMFM); (c) Multi-Scale Selection Module (MSSM).
    }
    \label{fig:zoomin}
\end{figure*}

As illustrated in Figure~\ref{fig:zoomin}, we provide model architecture details of three components: Intra-Object Relational Modeling (IORM), Cross-Modal Fusion Module (CMFM), and Multi-Scale Selection Module (MSSM), each corresponding to Figure~\ref{fig:zoomin}(a)--(c).

\paragraph{(a) Intra-Object Relational Modeling (IORM):}
Given multi-scale geometric features $\mathbf{F}_P \in \mathbb{R}^{C \times N}$, we first compute query, key, and value projections for each region:
\begin{align}
    \mathbf{q}_i &= W_q \mathbf{f}_i, \\
    \mathbf{k}_j &= W_k \mathbf{f}_j, \\
    \mathbf{v}_j &= W_v \mathbf{f}_j,
\end{align}
where $W_q, W_k, W_v \in \mathbb{R}^{C \times C}$ are learnable parameters and $\mathbf{f}_i$ denotes the feature of region $i$.
For each region $i$, we compute the scaled dot-product similarity to all other regions:
\begin{equation}
    \mathbf{S}_{ij} = \frac{\mathbf{q}_i^\top \mathbf{k}_j}{\sqrt{C}}.
\end{equation}
We then select the top-$k$ regions with the highest similarity scores and aggregate their corresponding value features to update the representation of region $i$:
\begin{equation}
    \tilde{\mathbf{f}}_i = \sum_{j \in \mathcal{N}_k(i)} \text{softmax}_j(\mathbf{S}_{ij}) \cdot \mathbf{v}_j.
\end{equation}
This procedure is applied independently at both large and small scales, producing enhanced features $\tilde{\mathbf{F}}_P^l\in \mathbb{R}^{C \times N_P^l}$ and $\tilde{\mathbf{F}}_P^s\in \mathbb{R}^{C \times N_P^s}$.

\paragraph{(b) Cross-Modal Fusion Module (CMFM):}
As shown in Figure~\ref{fig:zoomin}(b), let $\mathbf{F}_{\text{T}} \in \mathbb{R}^{C \times L}$ be the token-level language features and $\mathbf{F}_P \in \mathbb{R}^{C \times N}$ the geometric features (from IORM).  
\textit{Stage 1 (Text-to-Point Fusion):} For each scale, geometric features are queries and text features are keys/values:
\begin{equation}
    \mathbf{M}_P = \text{MHAttn}(\mathbf{F}_P, \mathbf{F}_{\text{T}}, \mathbf{F}_{\text{T}})
\end{equation}
\textit{Stage 2 (Point-to-Text Fusion):} Text features are updated by attending to geometry:
\begin{equation}
    \mathbf{M}_T = \text{MHAttn}(\mathbf{F}_{\text{T}}, \mathbf{M}_P, \mathbf{M}_P)
\end{equation}
Each $\text{MHAttn}$ denotes a multi-head cross-attention operation, and both stages are repeated at large and small scales.

\paragraph{(c) Multi-Scale Selection Module (MSSM):}
As illustrated in Figure~\ref{fig:zoomin}(c), after upsampling and modulation, we obtain features $\hat{\mathbf{G}}^l$ (large scale) and $\hat{\mathbf{G}}^s$ (small scale), both in $\mathbb{R}^{C \times N}$.
Adaptive scale fusion is performed via gating networks that predict soft weights $\alpha_l, \alpha_s$:
\begin{equation}
    \mathbf{F}_{\text{fuse}} = \alpha_l \hat{\mathbf{G}}^l + \alpha_s \hat{\mathbf{G}}^s, \quad \text{where}\ \alpha_l + \alpha_s = 1
\end{equation}
The gating network is a lightweight MLP applied to pooled features from each scale.

These formulations match the implementations in the main text, and the diagrams in Figure~\ref{fig:zoomin} provide a structural view of the data flow and operation within each module.

\subsection{Details of Our Hybrid Mask Loss}

For affordance mask prediction, we employ a hybrid loss function combining the focal loss and symmetric Dice loss to effectively address class imbalance and ensure robust mask predictions. Let $\mathbf{p} \in [0, 1]^{N \times C}$ denote the predicted mask probabilities, and $\mathbf{y} \in \{0, 1\}^{N \times C}$ the ground-truth binary mask, where $N$ is the number of points and $C$ is the number of classes.

\paragraph{Focal Loss}
The focal loss helps to focus on hard-to-classify examples by down-weighting the contribution of well-classified points. The equation for the focal loss is defined as:

\begin{equation}
\begin{split}
\mathcal{L}_{\mathrm{focal}} = & - (1-\alpha) \sum_{i=1}^N \sum_{c=1}^C (1 - y_{ic})\, p_{ic}^{\gamma} \log(1 - p_{ic} + \epsilon) \\
& - \alpha \sum_{i=1}^N \sum_{c=1}^C y_{ic}\, (1 - p_{ic})^{\gamma} \log(p_{ic} + \epsilon)
\end{split}
\end{equation}

Here, $\alpha$ is the weighting factor for the two classes, $\gamma$ is the focusing parameter, and $\epsilon$ is a small constant to prevent numerical instability.

\paragraph{Symmetric Dice Loss}
The symmetric Dice loss is used to measure the overlap between the predicted and ground-truth masks. It is split into positive and negative Dice components to account for both foreground and background classes. The Dice score for the positive and negative classes are given by:

\begin{equation}
\mathrm{Dice}_\text{pos}^{(c)} = \frac{\sum_{i=1}^N p_{ic} y_{ic} + \epsilon}{\sum_{i=1}^N |p_{ic}| + |y_{ic}| + \epsilon}
\end{equation}

\begin{equation}
\mathrm{Dice}_\text{neg}^{(c)} = \frac{\sum_{i=1}^N (1 - p_{ic})(1 - y_{ic}) + \epsilon}{\sum_{i=1}^N 2 - |p_{ic}| - |y_{ic}| + \epsilon}
\end{equation}

The total Dice loss is the sum of these two components across all classes, ensuring that both positive and negative classes are considered:

\begin{equation}
\mathcal{L}_{\mathrm{dice}} = \sum_{c=1}^C \left[ 1.5 - \mathrm{Dice}_\text{pos}^{(c)} - \mathrm{Dice}_\text{neg}^{(c)} \right]
\end{equation}

\paragraph{Total Hybrid Mask Loss}
The total hybrid mask loss combines the focal loss and Dice loss as follows:

\begin{equation}
\mathcal{L}_{\mathrm{mask}} = \mathcal{L}_{\mathrm{focal}} + \mathcal{L}_{\mathrm{dice}}
\end{equation}

In this setup, $\alpha = 0.25$, $\gamma = 2$, and $\epsilon = 10^{-6}$ by default.

This formulation encourages the model to focus on challenging examples and ensures accurate segmentation for both foreground and background regions.

\subsection{Feature dimensions}

\begin{table*}[t]
\centering
\caption{Dimensions and semantic roles of principal tensors in our method. The prototype set $\mathbf{A}$ is dynamic, initialized as $K=17$, and expands as new affordances are encountered during training.}
\label{tab:tensor-dims}
\begin{tabular}{lll}
\toprule
\textbf{Symbol} & \textbf{Shape} & \textbf{Description} \\
\midrule
$P$ & $2048 \times 3$& Input point cloud \\
$\mathbf{F}_{\text{T}}$ & $512 \times 40$& Token-level features of the instruction \\
$T_i$ & $512$& Embedding of part-specific token \\
$\mathbf{F}_P^l$ & $512 \times 64$& Large-scale geometric features \\
$\tilde{\mathbf{F}}_P^l$ & $512 \times 64$& Enhanced large-scale features \\
$\tilde{\mathbf{F}}_P^s$ & $512 \times 128$& Enhanced small-scale features \\
$\mathbf{M}_P^{l/s}$ & $512 \times 128$& Fused geometric features from cross-modal attention \\
$\mathbf{M}_T^{l/s}$ & $512 \times 40$& Structure-aware semantic features from cross-modal attention \\
$\hat{\mathbf{G}}^{l/s}$ & $512 \times 2048$& Modulated and upsampled features at each scale \\
$\mathbf{F}_{\text{fuse}}$ & $512 \times 2048$& Final fused representation for mask prediction \\
$\mathbf{A}$ & $K \times 512$& Prototype set for affordance concepts; $K$ dynamic, initialized as $17$ \\
$\mathbf{G}_{\text{gt}}$ & $512$& Ground-truth region embedding for semantic alignment \\
\bottomrule
\end{tabular}
\end{table*}

A concise summary of all major tensors and their dimensions is provided in Table~\ref{tab:tensor-dims}.

We summarize the principal tensors in our framework, clarifying their concrete dimensions and semantic functions as used in the model.

The input point cloud is represented as $P \in \mathbb{R}^{2048 \times 3}$, where $2048$ is the number of sampled points per object and each point has three spatial coordinates. The structured natural language instruction is encoded by a pretrained text encoder, yielding token-level features $\mathbf{F}_{\text{T}} \in \mathbb{R}^{512 \times 40}$, with a feature dimension of $512$ and a token sequence length of $40$. For explicit semantic–geometry alignment, a part-specific token (such as ``strap'') is extracted from the instruction, and its embedding $T_i \in \mathbb{R}^{512}$ is used as the semantic anchor for the corresponding object part.

The point cloud is partitioned into multi-scale regions. The large-scale geometric features are denoted by $\mathbf{F}_P^l \in \mathbb{R}^{512 \times 64}$, and after intra-object relational modeling, the enhanced features are $\tilde{\mathbf{F}}_P^l \in \mathbb{R}^{512 \times 64}$. At the small scale, the enhanced features are $\tilde{\mathbf{F}}_P^s \in \mathbb{R}^{512 \times 128}$, where $128$ is the number of small-scale regions.

Cross-modal fusion at each scale is performed via multi-head attention, yielding geometric features $\mathbf{M}_P^{l/s} \in \mathbb{R}^{512 \times 128}$ and structure-aware semantic features $\mathbf{M}_T^{l/s} \in \mathbb{R}^{512 \times 40}$. These features are modulated and propagated through a feature propagation module to produce $\hat{\mathbf{G}}^{l/s} \in \mathbb{R}^{512 \times 2048}$, aligning back to the original point cloud resolution.

The final fused representation for mask prediction is given by $\mathbf{F}_{\text{fuse}} \in \mathbb{R}^{512 \times 2048}$,  this fused tensor is used for both per-point affordance mask prediction and prototype-based abstraction.

Affordance-level reasoning is achieved through a dynamic prototype set $\mathbf{A} \in \mathbb{R}^{K \times 512}$, where each row $\mathbf{a}_k$ represents a canonical affordance concept in the feature space. Notably, the number of prototypes $K$ is not fixed a priori, but is dynamically expanded during training as new affordance categories are discovered from the natural language instructions. This enables the framework to avoid predefining a fixed set of affordance types and flexibly generalize to novel or previously unseen instructions. In practice, $K$ is initialized to $17$, corresponding to the default set of observed affordances in our setting, and will grow if new affordances emerge during training. The affordance semantics associated with each prototype are directly parsed from the instruction text.

For semantic alignment supervision, the ground-truth region embedding $\mathbf{G}_{\text{gt}} \in \mathbb{R}^{512}$ is computed via masked average pooling over $\mathbf{F}_{\text{fuse}}$, weighted by the binary ground-truth mask for each affordance.

\section{Baseline Implementation Details}
\label{baseline}

To ensure a fair and consistent comparison, we adapt all baseline methods to our experimental setting, applying minimal modifications unless otherwise required by modality or output format differences. The following summarizes the main adaptation strategies for each baseline; all other implementation aspects are preserved as much as possible.
\begin{itemize}
    \item XMF: Originally designed for image–point cloud fusion in completion tasks, we replace the image backbone in XMF with the same RoBERTa backbone used in our method, while keeping the cross-modal fusion structure unchanged. The final output head is adapted to produce affordance masks suitable for our task.
    \item IAG: As IAG is fundamentally image-driven, we substitute its image backbone with a RoBERTa-based text encoder to enable language input, with all other components and training configurations kept consistent with the original implementation.
    \item LASO: The LASO baseline shares the same task definition as ours; thus, we use its original design without any modification.
    \item OpenAD: Since OpenAD is a label-based 3D affordance grounding approach, we provide our input directly to the model and only adapt its output head to produce affordance masks aligned with our evaluation protocol.
    \item 3D-AffordanceLLM: This method closely matches our task. The only change is to remove the branch requiring an explicit answer input (absent in our data), and to adapt the output mask prediction to match our format; all other architectural details follow the original design.
\end{itemize}
We avoid any additional tuning or architectural changes not strictly necessary for task or modality adaptation, in order to preserve the integrity of each baseline.

\section{Evaluation Metrics}
\label{evaluation metrics}
Below are four used evaluation metrics and their calculation methods:

\begin{table*}[t]
\small
\centering
\begin{threeparttable}
\caption{Statistics of the proposed dataset under four evaluation settings.
PC–AFD: point cloud–affordance; Qst. Inst.: question instance; I/C: instance/category counts.}
\label{tab:split-statistics}

\begin{tabular}{l l r r r r}
\toprule
\textbf{Evaluation Setting} & \textbf{Split} &
\textbf{\#PC--AFD Pairs} & \textbf{\#Qst. Inst.} &
\textbf{\#OV--AFD (I/C)} & \textbf{\#Unseen OBJ (I/C)} \\
\midrule

\multirow{2}{*}{Closed-set Seen}
  & \g{Train} & \g{32\,709} & \g{742} & \g{\NA} & \g{\NA} \\
  & \g{Test}  & \g{4\,645}  & \g{53}  & \g{\NA} & \g{\NA} \\

\multirow{2}{*}{Closed-set Unseen}
  & Train & 29\,754 & 560 & \NA & \NA \\
  & Test  & 4\,645  & 53& \NA & 414 / 6 \\

\multirow{2}{*}{Open-set Full-view}
  & \g{Train} & \g{32\,709} & \g{742} & \g{\NA} & \g{\NA} \\
  & \g{Test}  & \g{4\,645}  & \g{69}  & \g{975 / 16} & \g{\NA} \\

\multirow{2}{*}{Open-set Partial-view}
  & Train & 130\,836 & 742 & \NA & \NA \\
  & Test  & 18\,580  & 69  & 3\,900 / 16 & \NA \\
\bottomrule
\end{tabular}
\end{threeparttable}
\end{table*}

\begin{enumerate}
    \item \textbf{AUC (Area Under Curve)}: \\
    AUC is an evaluation metric for point cloud saliency maps. It represents the area under the ROC (Receiver Operating Characteristic) curve. When predicting saliency maps, AUC treats the saliency map as a binary classifier under different thresholds. By measuring the true positive rate (TPR) and false positive rate (FPR) at each threshold, the ROC curve is plotted, and its area is calculated. A larger AUC value indicates higher accuracy of the model in predicting salient regions.
    
    \item \textbf{aIoU (Average Intersection Over Union)}: \\
    aIoU  is used to evaluate the overlap between the predicted affordance region and the ground truth region in the point cloud. IoU measures the similarity by calculating the ratio of the intersection to the union of the predicted and ground truth regions. The formula is as follows:
    \begin{equation}
        \text{IoU} = \frac{\text{TP}}{\text{TP} + \text{FP} + \text{FN}},
    \end{equation}
    where \(\text{TP}\) (True Positive) represents the number of correctly predicted positive examples, \(\text{FP}\) (False Positive) represents the number of incorrectly predicted positive examples, and \(\text{FN}\) (False Negative) represents the number of missed positive examples. A higher IoU value indicates a greater similarity between the predicted and ground truth regions.
    
    \item \textbf{SIM (Similarity)}: \\
    SIM measures the similarity between the predicted saliency map and the ground truth saliency map. For a predicted map \(P\) and a ground truth map \(Q_D\), SIM is computed by taking the minimum value between them at each pixel position and summing these minimum values. Both the predicted map and the ground truth map need to be normalized so that their pixel sums equal 1. The formula for SIM is:
    \begin{equation}
        \text{SIM}(P, Q_D) = \sum_i \min(P_i, Q_{D_i}).
    \end{equation}
    A higher SIM value indicates greater similarity between the prediction and the ground truth.
    
    \item \textbf{MAE (Mean Absolute Error)}: \\
    MAE evaluates the point-wise error between the predicted saliency map and the ground truth saliency map. It is calculated by taking the sum of the absolute values of all prediction errors and dividing it by the total number of points \(n\):
    \begin{equation}
        \text{MAE} = \frac{1}{n} \sum_{i=1}^n |e_i|,
    \end{equation}
    where \(e_i\) is the error for each point. A smaller MAE value indicates a smaller difference between the prediction and the ground truth.
\end{enumerate}

\section{Training Details}
\label{training}

All models are implemented in PyTorch and trained on a single NVIDIA H20 GPU. For both OpenAfford and LASO datasets, we use the Adam optimizer with $\beta_1 = 0.9$, $\beta_2 = 0.999$, $\epsilon=10^{-8}$, and weight decay of $5\times 10^{-4}$. The learning rate is scheduled via cosine annealing, with an initial value of $6\times 10^{-4}$ for OpenAfford and $1\times 10^{-4}$ for LASO, following practices in prior work.

The batch size is set to $96$ for OpenAfford and $32$ for LASO, as specified in the main text. All models are trained for $120$ epochs on OpenAfford, and $50$ epochs on LASO to match the protocol of the original benchmark.

Training on LASO follows the official data partitions provided by the benchmark. For our proposed OpenAfford dataset, training and validation splits are constructed as described in the main text. For all experiments, we report the evaluation metrics (aIoU, AUC, SIM, and MAE) corresponding to the epoch achieving the highest validation aIoU. Random seed is fixed to 42 for all runs to ensure reproducibility.

During training, input point clouds are uniformly sampled to 2048 points per shape, and input text is tokenized to a maximum of 40 tokens. Data loading leverages the PyTorch DataLoader with four worker threads for training and single-threaded validation.

All models are trained from scratch unless otherwise specified; resume from checkpoints is optionally supported. Model and optimizer states are periodically saved for reproducibility and ablation.

All remaining details follow the defaults provided in the released code and config files. For OpenAfford, all models (including baselines) are retrained for 120 epochs under the same protocol for fair comparison. For LASO, the training epoch is set to 50 to match the official implementation.

\section{Ablation Study Details}
\label{ablation}

For all ablation experiments, we follow the same training and evaluation protocols as the main model, with modifications applied only to the targeted components. All hyperparameters and training settings are kept identical across ablations to ensure fair comparison.

\begin{itemize}
    \item \textbf{PIG (Part-aware Instruction Generation):} Ablating PIG entails removing the structured language transformation and its downstream effects, which consequently disables explicit part-aware alignment and prototype association.
    \item \textbf{PSGA (Part-aware Semantic–Geometric Alignment):} In this variant, all branches responsible for extracting ground-truth region geometric features are removed, along with the corresponding loss terms.
    \item \textbf{IORM (Intra-Object Relational Modeling):} The IORM module is omitted entirely from the tensor flow, with all other architectural and training details left unchanged.
    \item \textbf{APA (Affordance Prototype Aggregation):} Removing APA eliminates all prototype-related parameters and loss functions, as well as all branches that compute geometric features for predicted regions.
\end{itemize}

In each ablation, the model architecture is minimally adapted to preserve the remaining components, and all experiments are conducted under consistent data and optimization settings for comparability.

\section{Experiments}
\label{experiments}
\subsection{Dataset Details}

We provide a comprehensive benchmark for language-driven 3D affordance grounding, with detailed statistics for each evaluation split presented in Table~\ref{tab:split-statistics}. The dataset covers both closed-set and open-set scenarios, supporting systematic assessment of semantic, categorical, and geometric generalization. All training and test splits are constructed to facilitate rigorous zero-shot evaluation on open-vocabulary queries.
While open-vocabulary (OV) affordance queries are created for both training and test splits, we emphasize that only the test set OV queries are used for zero-shot evaluation in this work. The training split with OV queries is made available primarily to encourage future research on multi-label or generalized affordance understanding.

\begin{table}[t]
\centering
\caption{Impact of part focus ambiguity rate on open-set full-view performance. Artificial ambiguity rates are introduced to test model robustness. OBJ denotes object.}
\label{tab:llm-part-error}
\begin{tabular}{ccccc}
\toprule
\textbf{OBJ Error Rate}& \textbf{aIoU~$\uparrow$} & \textbf{AUC~$\uparrow$} & \textbf{SIM~$\uparrow$} & \textbf{MAE~$\downarrow$} \\
\midrule
0\%   &             18.38&             82.33&             0.606&             0.110\\
10\%  &             17.43&             82.57&             0.603&             0.110\\
20\%  &             17.41&             82.57&             0.603&             0.111\\
\bottomrule
\end{tabular}
\end{table}

\begin{table}[t]
\centering
\caption{Impact of affordance misclassification rate on open-set full-view performance. Artificial error rates are introduced for robustness analysis. AFD denotes affordance.}
\label{tab:llm-affordance-error}
\begin{tabular}{ccccc}
\toprule
\textbf{AFD Error Rate}& \textbf{aIoU~$\uparrow$} & \textbf{AUC~$\uparrow$} & \textbf{SIM~$\uparrow$} & \textbf{MAE~$\downarrow$} \\
\midrule
0\%   &             18.38&             82.33&             0.606&             0.110\\
10\%&             17.22&             82.25&             0.600&             0.111\\
20\%&             17.03&             81.46&             0.587&             0.113\\
\bottomrule
\end{tabular}
\end{table}

\begin{figure*}[t]
    \centering
    \includegraphics[width=1\linewidth]{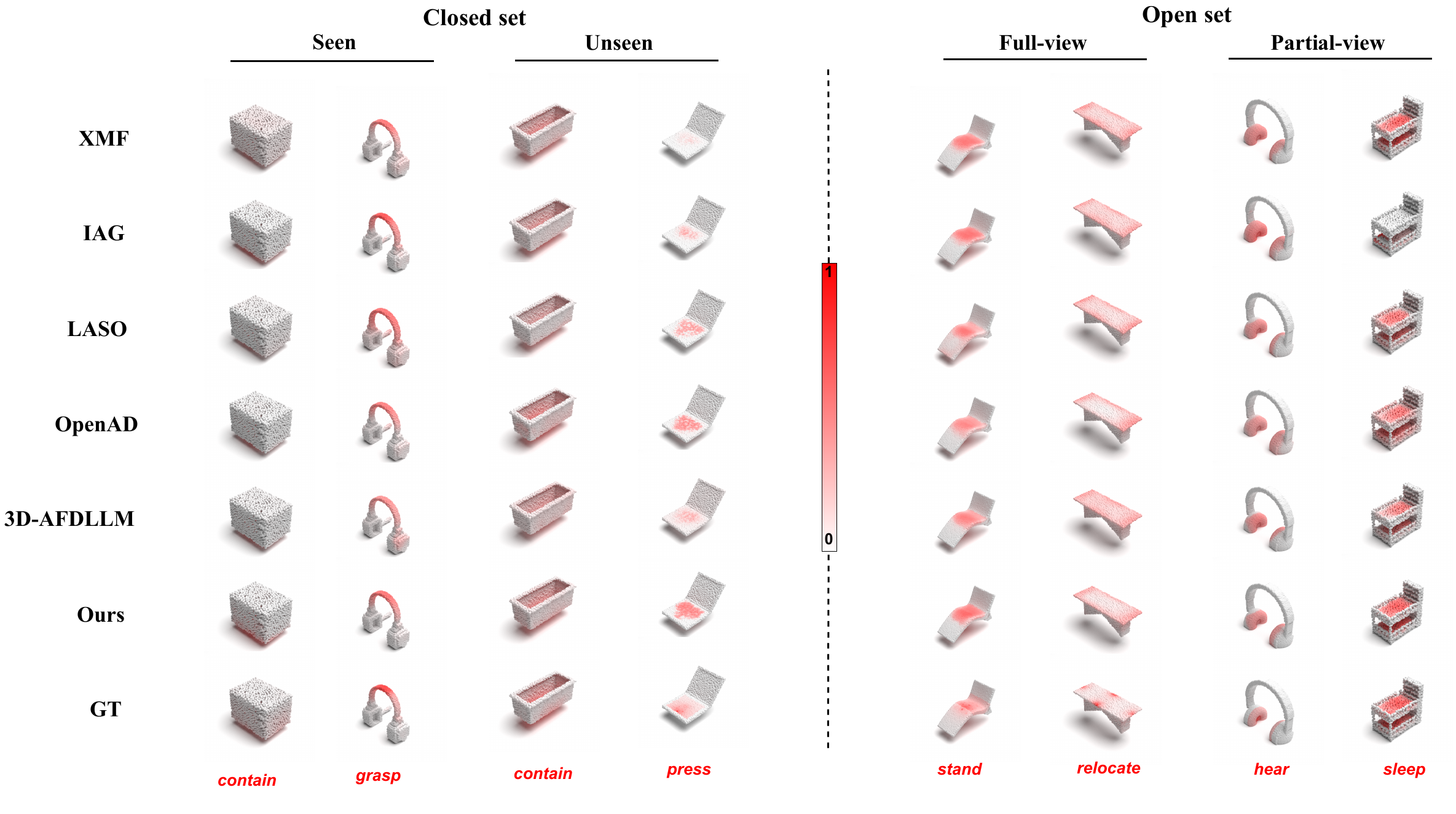}
    \caption{Additional qualitative visualizations of affordance grounding results across four evaluation settings.}
    \label{fig:mva}
\end{figure*}

\begin{figure}[t]
    \centering
    \includegraphics[width=1\linewidth]{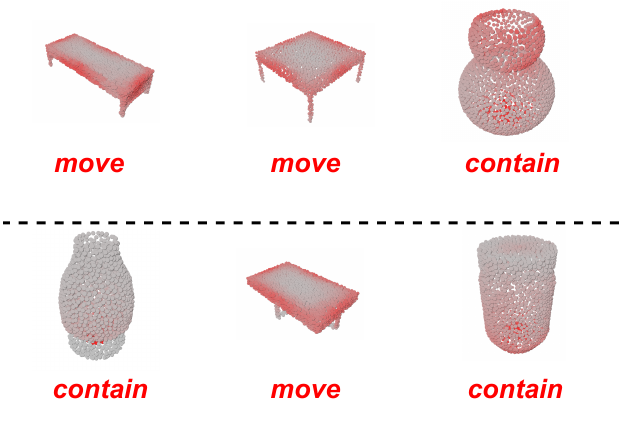}
    \caption{More cross-dataset zero-shot predictions on ShapeNet objects, illustrating semantic consistency and domain robustness.}
    \label{fig:cva}
\end{figure}

\begin{figure}[t]
    \centering
    \includegraphics[width=1\linewidth]{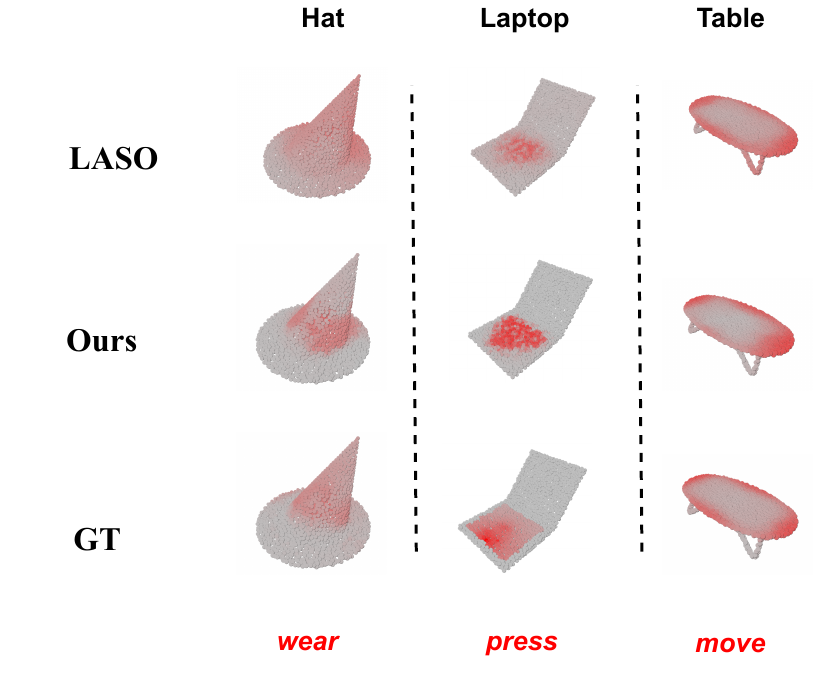}
    \caption{Qualitative visualizations of affordance grounding results on LASO objects.}
    \label{fig:laso}
\end{figure}

\begin{figure}[t]
    \centering
    \includegraphics[width=1\linewidth]{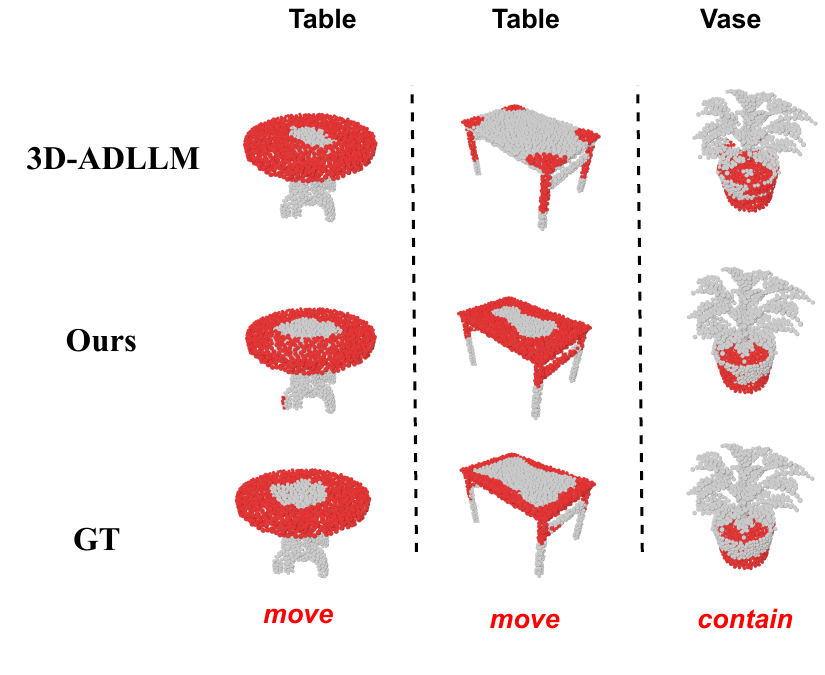}
    \caption{Qualitative visualizations of affordance grounding results on IRAS objects.}
    \label{fig:iras}
\end{figure}

\subsection{LLM-Related Latency and Error Analysis}

Our pipeline relies on a large language model (LLM) for transforming raw natural language questions into structured, part-aware instructions. While this enables robust semantic parsing, it also introduces two main limitations: inference latency and occasional misinterpretation of user queries.

\subsubsection{Latency Analysis.}
Using Mistral-7B as the instruction generator, we observe an average processing time of approximately 0.3–0.5 seconds per question on a single NVIDIA H20 GPU for short instructions. In comparison, the total model inference time without LLM transformation is 0.694 seconds per sample. While the LLM-based instruction generation constitutes a notable portion of the overall inference latency, it remains practical for most research and prototyping scenarios. For further deployment in latency-sensitive or large-scale settings, future work may explore fine-tuning lightweight language models or employing more efficient instruction generation strategies to further reduce this overhead.

\subsubsection{Error Analysis.}
Although our part-aware instruction generation module—powered by the LLM—achieves near-perfect accuracy on the OpenAfford benchmark, we conduct controlled experiments to assess our model's robustness under hypothetical misinterpretation scenarios. Specifically, we simulate two types of instruction errors and measure their effects on performance in the open-set full-view setting.

\paragraph{Affordance Misclassification.}
In our empirical evaluation, no genuine affordance-level misclassification errors were observed. To systematically evaluate model sensitivity, we artificially introduce affordance mislabeling by corrupting the part-aware instruction generation for 10\% and 20\% of the point cloud–affordance pairs during testing. As shown in Table~\ref{tab:llm-affordance-error}, increasing the misclassification rate results in a clear degradation of aIoU and other metrics, indicating that accurate affordance grounding is essential for reliable functional segmentation.

\begin{table}[t]
\centering
\caption{Impact of loss balancing coefficients $\beta_1$ and $\beta_2$ on model performance in the open-set full-view setting.}
\label{tab:hyper-sensitivity}
\begin{tabular}{cccccc}
\toprule
$\beta_1$ & $\beta_2$ & \textbf{aIoU~$\uparrow$} & \textbf{AUC~$\uparrow$} & \textbf{SIM~$\uparrow$} & \textbf{MAE~$\downarrow$} \\
\midrule
          0.2&           0.6&                 18.38&                82.33&                0.606&                  0.110\\
          0.3&           0.5&                 17.55&                82.30&                0.605&                  0.101\\
          0.5&           0.3&                 17.77&                82.78&                0.596&                  0.104\\
\bottomrule
\end{tabular}
\end{table}

\paragraph{Part Focus Ambiguity.}
Actual part focus misinterpretation by the LLM is extremely rare in our tests, typically limited to fine-grained distinctions between semantically related parts (e.g., ``mattress'' vs.\ ``surface'' for beds). To stress-test the model, we artificially introduce part focus errors by corrupting the part-aware instruction generation for 10\% and 20\% of the point cloud–affordance pairs during testing. As summarized in Table~\ref{tab:llm-part-error}, performance declines moderately with increasing ambiguity, but the overall drop is less severe than for affordance-level mistakes. This demonstrates the robustness of our method.

\subsection{Hyperparameter Sensitivity Analysis}
To evaluate the robustness of our framework to hyperparameter choices, we conduct a sensitivity analysis by varying the balancing coefficients $\beta_1$ and $\beta_2$ in the total loss. Experiments are performed under the open-set full-view evaluation setting on the OpenAfford benchmark. As summarized in Table~\ref{tab:hyper-sensitivity}, the performance metrics (aIoU, AUC, SIM, and MAE) remain stable across a wide range of hyperparameter values, demonstrating that our approach is not overly sensitive to reasonable variations in these loss balancing weights.

Overall, the results indicate that our model is robust to the selection of loss balancing weights, as illustrated in Table~\ref{tab:hyper-sensitivity}.

\subsection{Additional Visualizations}

We provide additional visualization results. Figure~\ref{fig:mva} presents more visualization examples under all four evaluation splits on OpenAfford, and Figure~\ref{fig:cva} shows more cross-dataset predictions. These results further demonstrate the model's prediction quality and cross-domain consistency. Further visualizations on LASO dataset and IRAS dataset are shown in Figures~\ref{fig:laso} and~\ref{fig:iras}, respectively, demonstrating the model's cross-domain consistency.


%% file: main.bbl
\begin{thebibliography}{49}
\providecommand{\natexlab}[1]{#1}
\providecommand{\url}[1]{\texttt{#1}}
\expandafter\ifx\csname urlstyle\endcsname\relax
  \providecommand{\doi}[1]{doi: #1}\else
  \providecommand{\doi}{doi: \begingroup \urlstyle{rm}\Url}\fi

\bibitem[Achlioptas et~al.(2020)Achlioptas, Abdelreheem, Xia, Elhoseiny, and Guibas]{Achlioptas2020}
Panos Achlioptas, Ahmed Abdelreheem, Fei Xia, Mohamed Elhoseiny, and Leonidas~J. Guibas.
\newblock {ReferIt3D: Neural Listeners for Fine-Grained 3D Object Identification in Real-World Scenes}.
\newblock In \emph{European Conference on Computer Vision (ECCV)}, 2020.

\bibitem[Aiello et~al.(2022)Aiello, Valsesia, and Magli]{aiello2022XMF}
Emanuele Aiello, Diego Valsesia, and Enrico Magli.
\newblock Cross‑modal learning for image‑guided point cloud shape completion.
\newblock In \emph{Advances in Neural Information Processing Systems}, 2022.

\bibitem[Azuma et~al.(2022)Azuma, Miyanishi, Kurita, and Kawanabe]{Azuma2022}
Daichi Azuma, Taiki Miyanishi, Shuhei Kurita, and Motoaki Kawanabe.
\newblock {ScanQA: 3D Question Answering for Spatial Scene Understanding}.
\newblock In \emph{Proceedings of the IEEE/CVF Conference on Computer Vision and Pattern Recognition (CVPR)}, 2022.

\bibitem[Chang et~al.(2015)Chang, Funkhouser, Guibas, Hanrahan, Huang, Li, Savarese, Savva, Song, Su, et~al.]{chang2015shapenet}
Angel~X Chang, Thomas Funkhouser, Leonidas Guibas, Pat Hanrahan, Qixing Huang, Zimo Li, Silvio Savarese, Manolis Savva, Shuran Song, Hao Su, et~al.
\newblock Shapenet: An information-rich 3d model repository.
\newblock \emph{arXiv preprint arXiv:1512.03012}, 2015.

\bibitem[Chen et~al.(2020)Chen, Chang, and Niessner]{Chen2020}
Dave~Zhenyu Chen, Angel~X. Chang, and Matthias Niessner.
\newblock {ScanRefer: 3D Object Localization in RGB-D Scans Using Natural Language}.
\newblock In \emph{European Conference on Computer Vision (ECCV)}, 2020.

\bibitem[Chen et~al.(2022)Chen, Guhur, Tapaswi, Schmid, and Laptev]{Chen2022LCSR}
Shizhe Chen, Pierre-Louis Guhur, Makarand Tapaswi, Cordelia Schmid, and Ivan Laptev.
\newblock Language conditioned spatial relation reasoning for 3d object grounding.
\newblock In \emph{Advances in Neural Information Processing Systems (NeurIPS)}, 2022.

\bibitem[Chu et~al.(2025)Chu, Deng, Lv, Chen, Li, Hao, and Nie]{Chu2025AffordanceLLM}
Hengshuo Chu, Xiang Deng, Qi Lv, Xiaoyang Chen, Yinchuan Li, Jianye Hao, and Liqiang Nie.
\newblock 3d‑affordancellm: Harnessing large language models for open‑vocabulary affordance detection in 3d worlds.
\newblock In \emph{ICLR, 2025}, 2025.

\bibitem[Deng et~al.(2021)Deng, Xu, Wu, Chen, and Jia]{Deng2021AffordanceNet}
Shengheng Deng, Xun Xu, Chaozheng Wu, Ke Chen, and Kui Jia.
\newblock 3d affordancenet: A benchmark for visual object affordance understanding.
\newblock In \emph{Proc. IEEE/CVF Conf. on Computer Vision and Pattern Recognition (CVPR)}, 2021.

\bibitem[Devlin et~al.(2019)Devlin, Chang, Lee, and Toutanova]{Devlin2019}
Jacob Devlin, Ming{-}Wei Chang, Kenton Lee, and Kristina Toutanova.
\newblock {BERT: Pre-training of Deep Bidirectional Transformers for Language Understanding}.
\newblock In \emph{Proceedings of the Conference of the North American Chapter of the Association for Computational Linguistics (NAACL)}, 2019.

\bibitem[Ding et~al.(2023)Ding, Yang, Xue, Zhang, Bai, and Qi]{Ding2023PLA}
Runyu Ding, Jihan Yang, Chuhui Xue, Wenqing Zhang, Song Bai, and Xiaojuan Qi.
\newblock {PLA}: Language-driven open-vocabulary 3d scene understanding.
\newblock In \emph{Proceedings of the IEEE/CVF Conference on Computer Vision and Pattern Recognition (CVPR)}, 2023.

\bibitem[Driess et~al.(2023)Driess, Xia, Sajjadi, Lynch, Chowdhery, Ichter, Wahid, Tompson, Vuong, Yu, et~al.]{Driess2023PaLM-E}
Danny Driess, Fei Xia, Mehdi~SM Sajjadi, Corey Lynch, Aakanksha Chowdhery, Brian Ichter, Ayzaan Wahid, Jonathan Tompson, Quan Vuong, Tianhe Yu, et~al.
\newblock Palm-e: an embodied multimodal language model.
\newblock In \emph{Proceedings of the 40th International Conference on Machine Learning}, pages 8469--8488, 2023.

\bibitem[Fan et~al.(2017)Fan, Cheng, Liu, Li, and Borji]{Fan2017StructureMeasure}
Deng{-}Ping Fan, Ming{-}Ming Cheng, Yun Liu, Tao Li, and Ali Borji.
\newblock Structure-measure: A new way to evaluate foreground maps.
\newblock In \emph{Proc.\ IEEE Int.\ Conf.\ on Computer Vision (ICCV)}, pages 4548--4557, 2017.

\bibitem[Gao et~al.(2025)Gao, Zhang, Qu, Wang, Wang, Ding, and Zhao]{Gao2025MIFAG}
Xianqiang Gao, Pingrui Zhang, Delin Qu, Dong Wang, Zhigang Wang, Yan Ding, and Bin Zhao.
\newblock Learning 2d invariant affordance knowledge for 3d affordance grounding.
\newblock In \emph{Proc. AAAI Conf. on Artificial Intelligence (AAAI)}, pages 3095--3103, 2025.

\bibitem[Gibson(2014)]{Gibson_2014}
JJ Gibson.
\newblock The ecological approach to visual perception: classic edition, 2014.

\bibitem[Grauman et~al.(2022)Grauman, Westbury, Byrne, Chavis, Furnari, Girdhar, Hamburger, Jiang, Liu, Liu, et~al.]{Gadre2022Ego4D}
Kristen Grauman, Andrew Westbury, Eugene Byrne, Zachary Chavis, Antonino Furnari, Rohit Girdhar, Jackson Hamburger, Hao Jiang, Miao Liu, Xingyu Liu, et~al.
\newblock Ego4d: Around the world in 3,000 hours of egocentric video.
\newblock In \emph{Proceedings of the IEEE/CVF conference on computer vision and pattern recognition}, pages 18995--19012, 2022.

\bibitem[Huang et~al.(2023)Huang, Mees, Zeng, and Burgard]{Huang2023VLMaps}
Chenguang Huang, Oier Mees, Andy Zeng, and Wolfram Burgard.
\newblock Visual language maps for robot navigation.
\newblock In \emph{Proc. IEEE Int. Conf. on Robotics and Automation (ICRA)}, page to appear, 2023.

\bibitem[Huang et~al.(2024)Huang, Wu, Chen, Zhao, Zhu, and Lasenby]{Huang2024OpenIns3D}
Zhening Huang, Xiaoyang Wu, Xi Chen, Hengshuang Zhao, Lei Zhu, and Joan Lasenby.
\newblock Openins3d: Snap and lookup for 3d open-vocabulary instance segmentation.
\newblock In \emph{Proc. European Conf. on Computer Vision (ECCV)}, 2024.

\bibitem[Jain et~al.(2022)Jain, Gkanatsios, Mediratta, and Fragkiadaki]{Jain2022}
Ayush Jain, Nikolaos Gkanatsios, Ishita Mediratta, and Katerina Fragkiadaki.
\newblock {Bottom-Up Top-Down Detection Transformers for Language Grounding in Images and Point Clouds}.
\newblock In \emph{European Conference on Computer Vision (ECCV)}, 2022.

\bibitem[Kerr et~al.(2023)Kerr, Kim, Goldberg, Kanazawa, and Tancik]{Kerr2023LERF}
Jason Kerr, Christopher~M. Kim, Ken Goldberg, Angjoo Kanazawa, and Matthew Tancik.
\newblock {LERF}: Language embedded radiance fields.
\newblock In \emph{Proceedings of the IEEE/CVF International Conference on Computer Vision (ICCV)}, pages 19729--19739, 2023.

\bibitem[Kingma and Ba(2015)]{kingma2015adam}
Diederik~P. Kingma and Jimmy Ba.
\newblock Adam: A method for stochastic optimization.
\newblock \emph{International Conference on Learning Representations (ICLR)}, 2015.

\bibitem[Li and Sigal(2021)]{Li_Sigal_2021ReferTrans}
Muchen Li and Leonid Sigal.
\newblock Referring transformer: A one-step approach to multi-task visual grounding.
\newblock \emph{Neural Information Processing Systems,Neural Information Processing Systems}, 2021.

\bibitem[Li et~al.(2024)Li, Zhao, Xiao, Feng, Wang, and Chua]{Li2024LASO}
Yicong Li, Na Zhao, Junbin Xiao, Chun Feng, Xiang Wang, and Tat-Seng Chua.
\newblock Laso: Language-guided affordance segmentation on 3d object.
\newblock In \emph{Proc. IEEE/CVF Conf. on Computer Vision and Pattern Recognition (CVPR)}, pages 14251--14260, 2024.

\bibitem[Lin et~al.(2017)Lin, Goyal, Girshick, He, and Doll{\'a}r]{lin2017focal}
Tsung-Yi Lin, Priya Goyal, Ross Girshick, Kaiming He, and Piotr Doll{\'a}r.
\newblock Focal loss for dense object detection.
\newblock In \emph{Proceedings of the IEEE international conference on computer vision}, pages 2980--2988, 2017.

\bibitem[Liu et~al.(2023{\natexlab{a}})Liu, Ding, and Jiang]{liu2023ReLA}
Chang Liu, Henghui Ding, and Xudong Jiang.
\newblock Gres: Generalized referring expression segmentation.
\newblock In \emph{Proceedings of the IEEE/CVF conference on computer vision and pattern recognition}, pages 23592--23601, 2023{\natexlab{a}}.

\bibitem[Liu et~al.(2023{\natexlab{b}})Liu, Shi, Kuang, Zhu, Li, Han, Cai, Porikli, and Su]{Liu2023}
Minghua Liu, Ruoxi Shi, Kaiming Kuang, Yinhao Zhu, Xuanlin Li, Shizhong Han, Hong Cai, Fatih Porikli, and Hao Su.
\newblock {OpenShape: Scaling Up 3D Shape Representation Towards Open-World Understanding}.
\newblock In \emph{Advances in Neural Information Processing Systems (NeurIPS)}, 2023{\natexlab{b}}.

\bibitem[Liu et~al.(2019)Liu, Ott, Goyal, Du, Joshi, Chen, Levy, Lewis, Zettlemoyer, and Stoyanov]{liu2019roberta}
Yinhan Liu, Myle Ott, Naman Goyal, Jingfei Du, Mandar Joshi, Danqi Chen, Omer Levy, Mike Lewis, Luke Zettlemoyer, and Veselin Stoyanov.
\newblock Roberta: A robustly optimized bert pretraining approach.
\newblock \emph{arXiv preprint arXiv:1907.11692}, 2019.

\bibitem[Lobo et~al.(2008)Lobo, Jiménez-Valverde, and Real]{Fawcett2006ROC}
Jorge~M. Lobo, Alberto Jiménez-Valverde, and Raimundo Real.
\newblock Auc: a misleading measure of the performance of predictive distribution models.
\newblock \emph{Global Ecology and Biogeography}, page 145–151, 2008.

\bibitem[Lu et~al.(2023)Lu, Xu, Wei, Xie, Tomizuka, Keutzer, and Zhang]{Lu2023OV3DET}
Yuheng Lu, Chenfeng Xu, Xiaobao Wei, Xiaodong Xie, Masayoshi Tomizuka, Kurt Keutzer, and Shanghang Zhang.
\newblock Open-vocabulary point-cloud object detection without 3d annotation.
\newblock In \emph{Proceedings of the IEEE/CVF Conference on Computer Vision and Pattern Recognition (CVPR)}, 2023.

\bibitem[Luo et~al.(2022)Luo, Fu, Kong, Gao, Ren, Shen, Xia, and Liu]{luo20223DSPS}
Junyu Luo, Jiahui Fu, Xianghao Kong, Chen Gao, Haibing Ren, Hao Shen, Huaxia Xia, and Si Liu.
\newblock 3d-sps: Single-stage 3d visual grounding via referred point progressive selection.
\newblock In \emph{Proceedings of the IEEE/CVF Conference on Computer Vision and Pattern Recognition}, pages 16454--16463, 2022.

\bibitem[Ma et~al.(2023)Ma, Yong, Zheng, Li, Liang, Zhu, and Huang]{Ma2023}
Xiaojian Ma, Silong Yong, Zilong Zheng, Qing Li, Yitao Liang, Song{-}Chun Zhu, and Siyuan Huang.
\newblock {SQA3D: Situated Question Answering in 3D Scenes}.
\newblock In \emph{International Conference on Learning Representations (ICLR)}, 2023.

\bibitem[Milletari et~al.(2016)Milletari, Navab, and Ahmadi]{milletari2016vnet}
Fausto Milletari, Nassir Navab, and Seyed-Ahmad Ahmadi.
\newblock V-net: Fully convolutional neural networks for volumetric medical image segmentation.
\newblock In \emph{2016 fourth international conference on 3D vision (3DV)}, pages 565--571. Ieee, 2016.

\bibitem[Nguyen et~al.(2023)Nguyen, Vu, Vuong, and Pham]{Nguyen2023OpenAD}
Tung Nguyen, Minh~Nhat Vu, Anh Vuong, and Quang-Cuong Pham.
\newblock Open-vocabulary affordance detection in 3d point clouds.
\newblock In \emph{Proceedings of the 2023 IEEE/RSJ International Conference on Intelligent Robots and Systems (IROS)}. IEEE, 2023.

\bibitem[Peng et~al.(2023)Peng, Genova, Jiang, Tagliasacchi, Pollefeys, and Funkhouser]{Peng2023OpenScene}
Songyou Peng, Kyle Genova, Chiyu~Max Jiang, Andrea Tagliasacchi, Marc Pollefeys, and Thomas Funkhouser.
\newblock Openscene: 3d scene understanding with open vocabularies.
\newblock In \emph{Proceedings of the IEEE/CVF Conference on Computer Vision and Pattern Recognition (CVPR)}, pages 815--824, 2023.

\bibitem[Qi et~al.(2017)Qi, Yi, Su, and Guibas]{qi2017pointnet++}
Charles~Ruizhongtai Qi, Li Yi, Hao Su, and Leonidas~J Guibas.
\newblock Pointnet++: Deep hierarchical feature learning on point sets in a metric space.
\newblock \emph{Advances in neural information processing systems}, 30, 2017.

\bibitem[Qi et~al.(2024)Qi, Dong, Zhang, Geng, Han, Ge, Yi, and Ma]{qi12shapellm}
Zekun Qi, Runpei Dong, Shaochen Zhang, Haoran Geng, Chunrui Han, Zheng Ge, Li Yi, and Kaisheng Ma.
\newblock Shapellm: Universal 3d object understanding for embodied interaction.
\newblock In \emph{European Conference on Computer Vision}, pages 214--238. Springer, 2024.

\bibitem[Qin et~al.(2024)Qin, Li, Zhou, Wang, and Pfister]{Qin2024LangSplat}
Minghan Qin, Wanhua Li, Jiawei Zhou, Haoqian Wang, and Hanspeter Pfister.
\newblock Langsplat: 3d language gaussian splatting.
\newblock In \emph{Proceedings of the IEEE/CVF Conference on Computer Vision and Pattern Recognition (CVPR)}, pages 20051--20060, 2024.

\bibitem[Tabib et~al.(2024)Tabib, Hegde, and Mudenagudi]{Tabib2024LGAffordNet}
Ramesh~Ashok Tabib, Dikshit Hegde, and Uma Mudenagudi.
\newblock Lg‑affordnet: A local geometry aware affordance detection network for 3d point clouds.
\newblock In \emph{CVPR Workshops (DLGC), 2024}, 2024.

\bibitem[Takmaz et~al.(2023)Takmaz, Fedele, Sumner, Pollefeys, Tombari, and Engelmann]{Takmaz2023OpenMask3D}
Ay\c{c}a Takmaz, Elisabetta Fedele, Robert~W. Sumner, Marc Pollefeys, Federico Tombari, and Francis Engelmann.
\newblock Openmask3d: Open-vocabulary 3d instance segmentation.
\newblock In \emph{Advances in Neural Information Processing Systems (NeurIPS)}, 2023.

\bibitem[Vobeck\'y et~al.(2023)Vobeck\'y, Sim\'eoni, Hurych, Gidaris, Bursuc, P\'erez, and Sivic]{Vobecky2023POP3D}
Anton\'in Vobeck\'y, Oriane Sim\'eoni, David Hurych, Spyros Gidaris, Andrei Bursuc, Patrick P\'erez, and Josef Sivic.
\newblock {POP-3D}: Open-vocabulary 3d occupancy prediction from images.
\newblock In \emph{Advances in Neural Information Processing Systems (NeurIPS)}, 2023.

\bibitem[Wan et~al.(2025)Wan, Gou, Liu, Zhu, and He]{obj_afford}
Xinhang Wan, Dongqiang Gou, Xinwang Liu, En Zhu, and Xuming He.
\newblock Object affordance recognition and grounding via multi-scale cross-modal representation learning.
\newblock \emph{arXiv preprint arXiv:2508.01184}, 2025.

\bibitem[Wang and Czarnecki(2025)]{Wang2025AIDE}
Yimu Wang and Krzysztof Czarnecki.
\newblock {AIDE}: Improving 3d open-vocabulary semantic segmentation by aligned vision-language learning.
\newblock In \emph{Proceedings of the IEEE/CVF Winter Conference on Applications of Computer Vision (WACV)}, 2025.

\bibitem[Xue et~al.(2023)Xue, Gao, Xing, Mart{\'{\i}}n-Mart{\'{\i}}n, Wu, Xiong, Xu, Niebles, and Savarese]{Xue2023}
Le Xue, Mingfei Gao, Chen Xing, Roberto Mart{\'{\i}}n-Mart{\'{\i}}n, Jiajun Wu, Caiming Xiong, Ran Xu, Juan~Carlos Niebles, and Silvio Savarese.
\newblock {ULIP: Learning a Unified Representation of Language, Images, and Point Clouds for 3D Understanding}.
\newblock In \emph{Proceedings of the IEEE/CVF Conference on Computer Vision and Pattern Recognition (CVPR)}, 2023.

\bibitem[Yamazaki et~al.(2023)Yamazaki, Hanyu, Vo, Pham, Tran, Doretto, Nguyen, and Le]{Yamazaki2023OpenFusion}
Kashu Yamazaki, Taisei Hanyu, Khoa Vo, Thang Pham, Minh Tran, Gianfranco Doretto, Anh Nguyen, and Ngan Le.
\newblock Open-fusion: Real-time open-vocabulary 3d mapping and queryable scene representation.
\newblock \emph{arXiv preprint arXiv:2310.03923}, 2023.

\bibitem[Yang et~al.(2024)Yang, Ju, and Yi]{Yang2024ImOV3D}
Timing Yang, Yuanliang Ju, and Li Yi.
\newblock {ImOV3D}: Learning open vocabulary point clouds 3d object detection from only 2d images.
\newblock In \emph{Advances in Neural Information Processing Systems (NeurIPS)}, 2024.

\bibitem[Yang et~al.(2023)Yang, Zhai, Luo, Cao, Luo, and Zha]{Yang2023IAG}
Yuhang Yang, Wei Zhai, Hongchen Luo, Yang Cao, Jiebo Luo, and Zheng-Jun Zha.
\newblock Grounding 3d object affordance from 2d interactions in images.
\newblock In \emph{Proc. IEEE/CVF Int. Conf. on Computer Vision (ICCV)}, pages 10905--10915, 2023.

\bibitem[Yu et~al.(2025)Yu, Wang, Shi, Luo, Yang, Yu, and Wang]{Yu2025SeqAfford}
Chunlin Yu, Hanqing Wang, Ye Shi, Haoyang Luo, Sibei Yang, Jingyi Yu, and Jingya Wang.
\newblock Seqafford: Sequential 3d affordance reasoning via multimodal large language model.
\newblock In \emph{Proc. IEEE/CVF Conf. on Computer Vision and Pattern Recognition (CVPR)}, pages 1691--1701, 2025.

\bibitem[Zhang et~al.(2021)Zhang, Yang, Constantinescu, Peng, M{\"u}ller, and Stiefelhagen]{Shan2023Trans4Trans}
Jiaming Zhang, Kailun Yang, Angela Constantinescu, Kunyu Peng, Karin M{\"u}ller, and Rainer Stiefelhagen.
\newblock Trans4trans: Efficient transformer for transparent object segmentation to help visually impaired people navigate in the real world.
\newblock In \emph{Proceedings of the IEEE/CVF International Conference on Computer Vision}, pages 1760--1770, 2021.

\bibitem[Zhao et~al.(2021)Zhao, Cai, Sheng, and Xu]{Zhao2021}
Lichen Zhao, Daigang Cai, Lu Sheng, and Dong Xu.
\newblock {3DVG-Transformer: Relation Modeling for Visual Grounding on Point Clouds}.
\newblock In \emph{Proceedings of the IEEE/CVF International Conference on Computer Vision (ICCV)}, 2021.

\bibitem[Zhu et~al.(2025)Zhu, Kong, Xu, Xia, Deng, Ye, Xiong, and Wang]{Zhu2025LMA}
He Zhu, Quyu Kong, Kechun Xu, Xunlong Xia, Bing Deng, Jieping Ye, Rong Xiong, and Yue Wang.
\newblock {Grounding 3D Object Affordance with Language Instructions, Visual Observations and Interactions}.
\newblock In \emph{Proceedings of the IEEE/CVF Conference on Computer Vision and Pattern Recognition (CVPR)}, 2025.

\end{thebibliography}
